\documentclass[10pt,twocolumn,letterpaper]{article}

\usepackage{iccv}              

\usepackage{multirow} 

\usepackage{caption}
\usepackage{comment}
\usepackage{etoc}
\usepackage{pifont}
\etocdepthtag.toc{mtchapter}
\etocsettagdepth{mtchapter}{subsection}
\etocsettagdepth{mtappendix}{none}

\definecolor{iccvblue}{rgb}{0.21,0.49,0.74}
\usepackage[pagebackref,breaklinks,colorlinks,allcolors=iccvblue]{hyperref}
\usepackage[capitalize]{cleveref}
\usepackage{dsfont}
\usepackage[linesnumbered,ruled,vlined]{algorithm2e}
\crefname{section}{Sec.}{Secs.}
\Crefname{section}{Section}{Sections}
\Crefname{table}{Table}{Tables}
\crefname{table}{Tab.}{Tabs.}

\def\paperID{4627} 
\def\confName{ICCV}
\def\confYear{2025}

\title{Generate, Refine, and Encode: Leveraging Synthesized Novel Samples \\ for On-the-Fly Fine-Grained Category Discovery}

\author{Xiao Liu$^{1,3}$\footnotemark[1]~, Nan Pu$^{2}$\footnotemark[1]~\footnotemark[2]~, Haiyang Zheng$^{2}$, Wenjing Li$^{1}$, Nicu Sebe$^{2}$, Zhun Zhong$^{1}$\footnotemark[2]~\\
$^{1}$Hefei University of Technology, 
~$^{2}$University of Trento, ~$^{3}$Helmholtz AI
}

\begin{document}
\maketitle
\begin{abstract}

In this paper, we investigate a practical yet challenging task: On-the-fly Category Discovery (OCD). This task focuses on the online identification of newly arriving stream data that may belong to both known and unknown categories, utilizing the category knowledge from only labeled data. Existing OCD methods are devoted to fully mining transferable knowledge from only labeled data. However, the transferability learned by these methods is limited because the knowledge contained in known categories is often insufficient, especially when few annotated data/categories are available in fine-grained recognition. To mitigate this limitation, we propose a diffusion-based OCD framework, dubbed Diff\textbf{GRE}, which integrates \textbf{G}eneration, \textbf{R}efinement, and \textbf{E}ncoding in a multi-stage fashion. Specifically, we first design an attribute-composition generation method based on cross-image interpolation in the diffusion latent space to synthesize novel samples. Then, we propose a diversity-driven refinement approach to select the synthesized images that differ from known categories for subsequent OCD model training. Finally, we leverage a semi-supervised leader encoding to inject additional category knowledge contained in synthesized data into the OCD models, which can benefit the discovery of both known and unknown categories during the on-the-fly inference process. Extensive experiments demonstrate the superiority of our DiffGRE over previous methods on six fine-grained datasets. The codes are available here: 
\href{https://github.com/XLiu443/DiffGRE}{https://github.com/XLiu443/DiffGRE}

\end{abstract}
\renewcommand{\thefootnote}{\fnsymbol{footnote}}
\footnotetext[1]{Equal contribution.}
\footnotetext[2]{Corresponding author.}
\vspace{-.2in}
\section{Introduction}
\label{sec:intro}

\begin{figure}[t]
  \centering
\includegraphics[width=\linewidth]{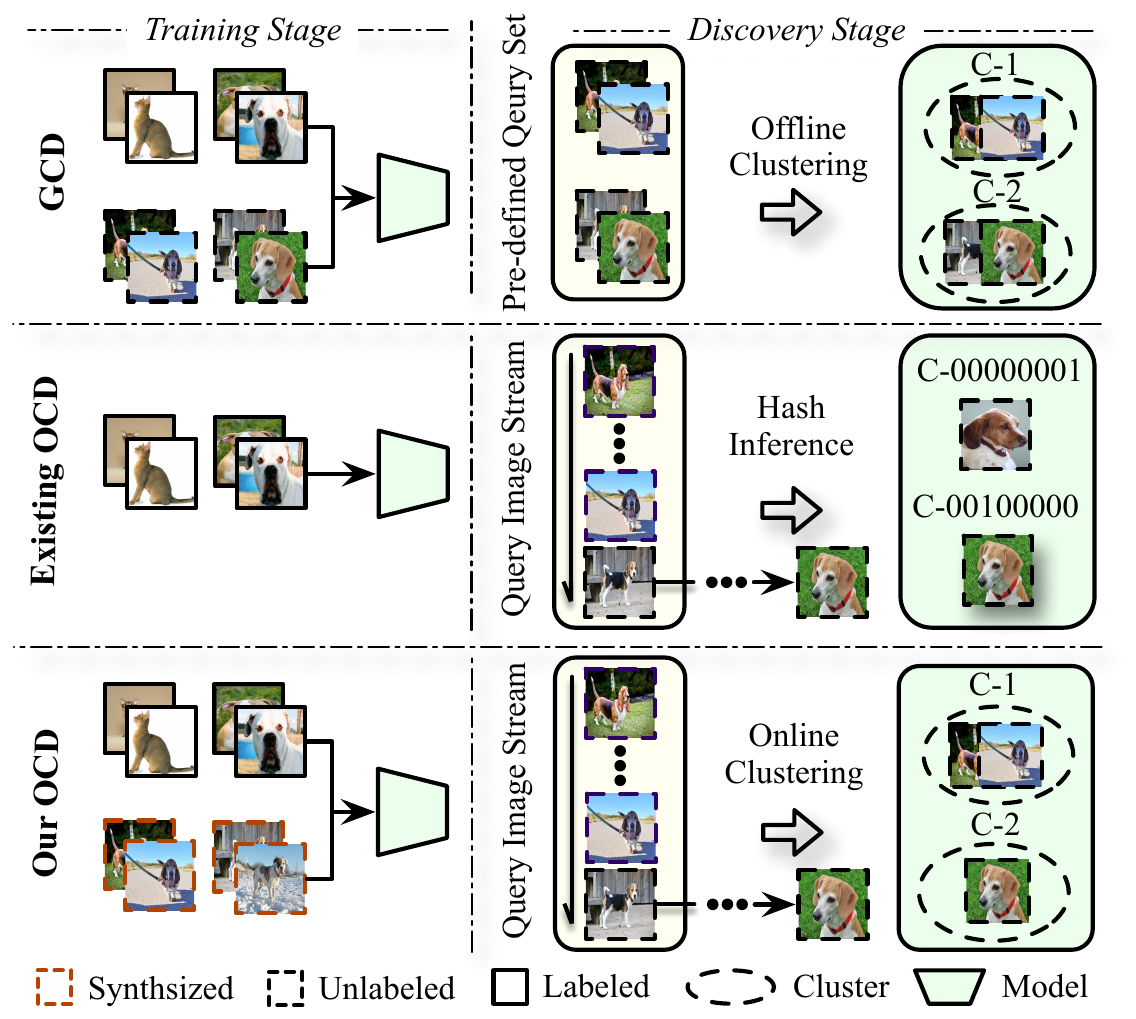}
    \caption{Comparison with GCD and OCD. Most existing GCD methods follow a transductive learning paradigm that requires previously accessing test data during training. Meanwhile, they utilize offline clustering to assign category labels for pre-collected query sets. In contrast, OCD models are trained with only labeled training data and can provide instance-wise inference for a stream of data that require discovery. Unlike existing OCD methods, our DiffGRE syntheses new training samples for OCD training, while performing more robust online cluster than hash inference.
    \label{fig:ocd_ncd}}
    \vspace{-1em}
\end{figure}

Generalized category discovery (GCD), an emerging and challenging problem, has garnered increasing attention due to its potential for widespread real-world applications, such as discovering new diseases~\cite{zhou2024novel}, bird species~\cite{pu2024federated}, and crops~\cite{Liu_2024_CVPR}. GCD aims to automatically cluster unlabeled data from unseen categories using knowledge from seen categories. However, current GCD methods~\cite{gcd,dccl,pu2024federated,zheng2024textual,yang2024learn} follow an offline inference paradigm, where category discovery is typically implemented by applying conventional clustering algorithms to a pre-collected batch of query data that requires discovery. This paradigm significantly limits the practicality of GCD techniques in real-world applications, where systems are expected to provide online feedback for each new instance.

\par
To address this limitation, On-the-fly Category Discovery (OCD)~\cite{ocd} is introduced to eliminate the requirement of accessing pre-defined query sets in training phrases and enables GCD models to generate instance feedback for streaming data, as shown in~\cref{fig:ocd_ncd}. 
Existing OCD methods that focus on learning transferable knowledge from only labeled data and categories are often inefficient, as the knowledge contained in known categories is often insufficient, especially in fine-grained scenarios where annotated data are often scarce. To address this limitation, we propose a diffusion-based OCD framework, namely DiffGRE, to synthesize new samples that contain additional category information and are helpful for fine-grained category discovery.

\par
Diff\textbf{GRE} mainly includes \textbf{G}eneration, \textbf{R}efinement, and \textbf{E}ncoding stages. Specifically, most existing diffusion-based data augmentation methods~\cite{he2022synthetic,feng2023diverse,wang2024enhance} either heavily rely on fine-tuning the pre-trained diffusion model on large-scale target data or aim at faithfulness augmentation (\eg, category-preserving generation~\cite{islam2024diffusemix, trabucco2023effective} or seen target-category argumentation~\cite{wang2024enhance}), which are not effective for the OCD task because 1) there has relatively few labeled data in OCD setting, especially for fine-grained datasets; 2) the data that require to discover include both known and unknown categories. However, we empirically find that augmenting only known-category data is struggling to benefit unknown-category discovery (see details in Appendix). 
\par
\begin{figure}[t]
  \centering
\includegraphics[width=\linewidth]{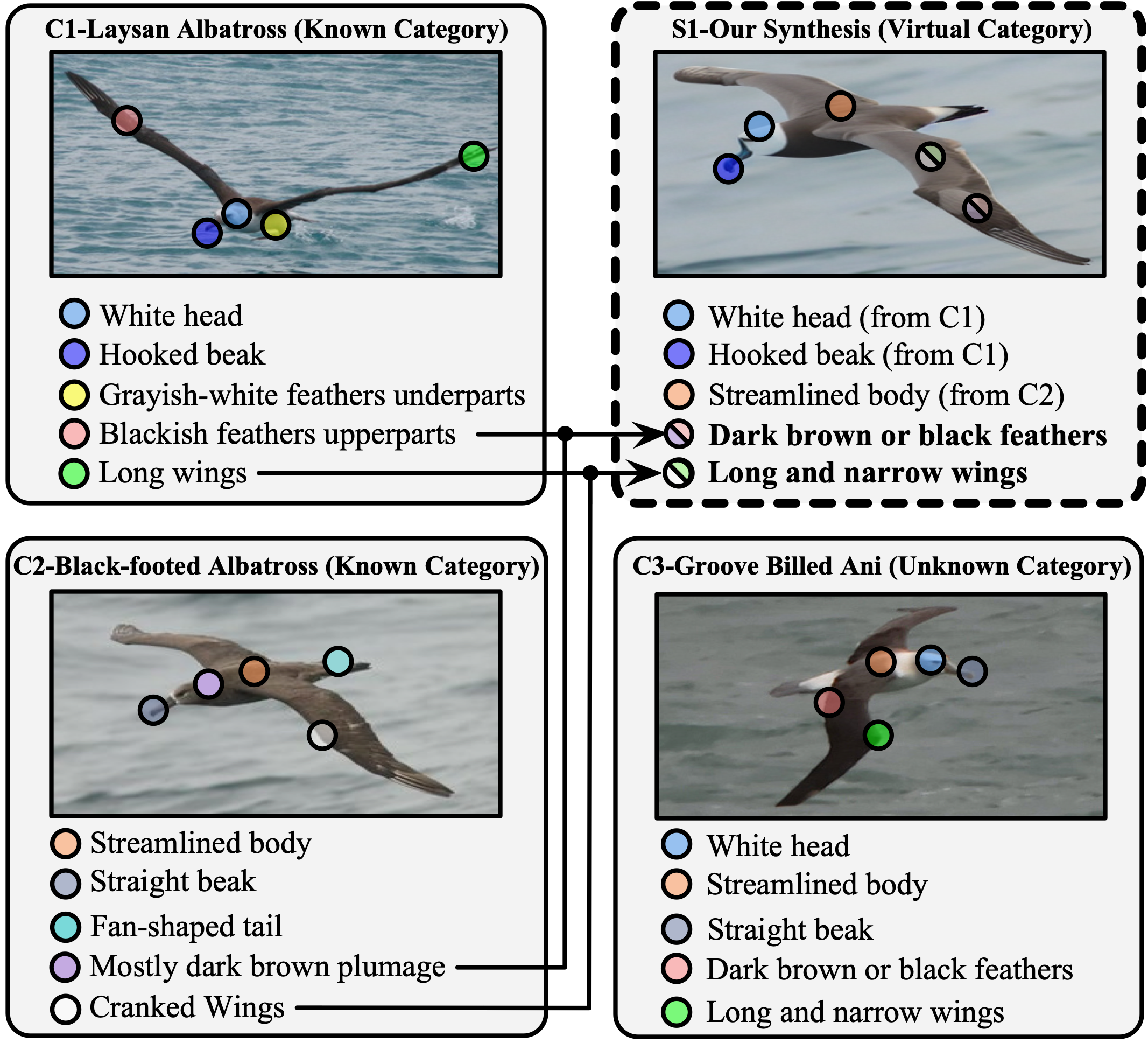}
    \caption{Motivation of this paper. In this case, we randomly sample two images from two known categories as inputs to synthesize a new image by attribute re-composition. We observe that the synthesized image has similar attributes to an unknown category defined in the OCD benchmark. This observation motivates us to fully leverage latent category attributes included in known categories to generate additional category information, thereby improving the existing OCD methods.
 \label{fig:motivation}}
    \vspace{-1em}
\end{figure}

Considering that 1) diffusion models already have a semantic latent space~\cite{DBLP:conf/iclr/KwonJU23,shuai2024latent, rombach2022high} and 2) unseen categories can be synthesized by composing existing semantic attributes in seen categories~\cite{li2024context, wang2023videocomposer, liu2024pmgnet}, we propose to explore a cross-category latent-space interpolation that can synthesize virtual categories through semantic Attribute Composition Generation (ACG). ACG employs the latent interpolation strategy, which not only interpolates diffusion latent features but also integrates interpolation on the CLIP~\cite{radford2021learning} embedding space. This enhances semantic information in the latent space, thereby enriching the attributes that can be combined. An example is shown in ~\cref{fig:motivation}, where two images are used as inputs for ACG, resulting in a synthesized image. We interpolate between a ``Laysan albatross'' image and a ``Black-footed albatross'' image and synthesize a new bird image that belongs to a virtual category. Surprisingly, by searching from both known- and unknown-category data, we find that the nearest neighbor of the generated image belongs to an unknown category, \eg, ``Groove-billed Ani'' in~\cref{fig:motivation}. ``Groove-billed Ani'' appears as a combination of the semantic attributes from ``Laysan albatross'' and ``Laysan albatross'', which shows the promising potential to synthesize unknown categories based on known-category attributes.

\par
Although the image synthesis approach can achieve semantic attribute re-composition, it cannot ensure that each generated image is likely a new-category image and is helpful for OCD due to the randomness of diffusion process~\cite{ho2020denoising}. To mitigate this side-effect, we design a lightweight Diversity-Driven Refinement (DDR) to select the generated images that include diverse category information as much as possible. DDR filters off synthesized images that are visually similar to known categories, according to the average similarities between the image features and category-level representations. This mechanism not only improves the diversity of augmented samples but also reduces the computational consumption for subsequent model training.
\par

To take full advantage of the refined synthesized data, we propose a Semi-supervised Leader Encoding (SLE) to assign reliable pseudo labels for them. Importantly, the encoded leader representations in SLE are further used to implement accurate on-the-fly inference based on the high-dimensional features outputted from the backbone network. This differs from the existing method~\cite{ocd} that employs hash code as category descriptors, which inevitably degrades the representation ability of the high-dimensional features due to the binary nature. In contrast, our SLE can directly operate on original features containing rich discriminative information following an online clustering paradigm, which hence leads to better OCD performance.

Our contribution can be summarized as follows:
\begin{itemize}
    \item We advocate embracing pre-trained and high-quality generative models based on the diffusion technique and introducing them into new category discovery tasks.
    \item We propose a new and plug-and-play DiffGRE framework to enhance the OCD abilities of existing methods.
    \item Extensive experiments verify the effectiveness of our DiffGRE on six fine-grained datasets.
\end{itemize}

\section{Related Work}
\label{sec:related_work}

\vspace{-.05in}

\textbf{Novel Category Discovery} (NCD), first introduced by DTC~\cite{dtc}, aims to categorize unlabeled novel classes by transferring knowledge from labeled known classes. However, existing NCD methods~\cite{dtc,ncl,uno} assume that all unlabeled data exclusively belong to novel classes. To release such an impractical assumption, Generalized Category Discovery (GCD) is proposed in~\cite{gcd}, allowing unlabeled data to be sampled from both novel and known classes. While existing NCD/GCD methods~\cite{ncl,uno,gcd,dccl,simgcd,pim,dpn-gcd,igcd,metagcd,proxy,simgcd} have shown promising results, two key assumptions still impede their real-world application. 1) These models heavily rely on a predetermined query set (the unlabeled dataset) during training, limiting their ability to handle truly novel samples and hindering generalization. 2) The offline batch processing of the query set during inference makes these models impractical for online scenarios where data emerges continuously and the model requires instant feedback. To address these limitations, Du et al. introduced the On-the-Fly Category Discovery (OCD)~\cite{ocd}, which removes the assumption of a predefined query set and requires instance feedback with stream data input. They proposed the SMILE method, which identifies the category of each instance via a hash-descriptor inference. 
\par\textit{Notably, compared with the above methods, our DiffGRE has two differences: 1) Unlike SMILE, which employs sensitive hash inference, DiffGRE directly leverages features for inference following an online clustering fashion. 2) Existing OCD methods utilize only labeled data for training. In contrast, our DiffGRE generates new samples and designs a new effective framework to exploit these synthesized data.}

\enlargethispage{4mm}
\par\noindent
\textbf{Data Synthetic with Diffusion Models}. Synthesizing new samples with diffusion models has been applied in various computer vision tasks, including image classification~\cite{tsai2025test,islam2024diffusemix, shipard2023diversity}, domain adaptation~\cite{du2023diffusion, kim2023podia, peng2024unsupervised} and zero/few-shot learning~\cite{clark2024text, you2024diffusion, li2023your}. Most of these methods aim to augment images to enhance diversity while keeping faithfulness, which can be broadly divided into two groups. The one~\cite{wang2024enhance, trabucco2023effective} is to generate images based on Img2Img diffusion with pre-defined or learned text prompts. The other leverages Latent Perturbation~\cite{fu2024dreamda, zhou2023training} to empower the ability of diffusion models to synthesize new samples. However, the former highly relies on prompt quality and usually can only generate new samples with high faithfulness and low diversity. The latter can synthesize images with high diversity, but usually, these samples contain massive noises. To mitigate these drawbacks, Diff-Mix~\cite{wang2024enhance} proposes to augment training images by transforming one known category to another known category in a mix-up manner. \par\textit{However, these methods focus on generating new images of known categories, which is relatively ineffective for OCD task that aims at new category discovery. To address this issue, we introduce DiffGRE to synthesize virtual category samples via attribute re-composition, tailor-made for OCD.}
\section{Method}
\label{sec:method}

\vspace{-.05in}

\begin{figure*}[!t]
  \centering
  \includegraphics[width=\linewidth]{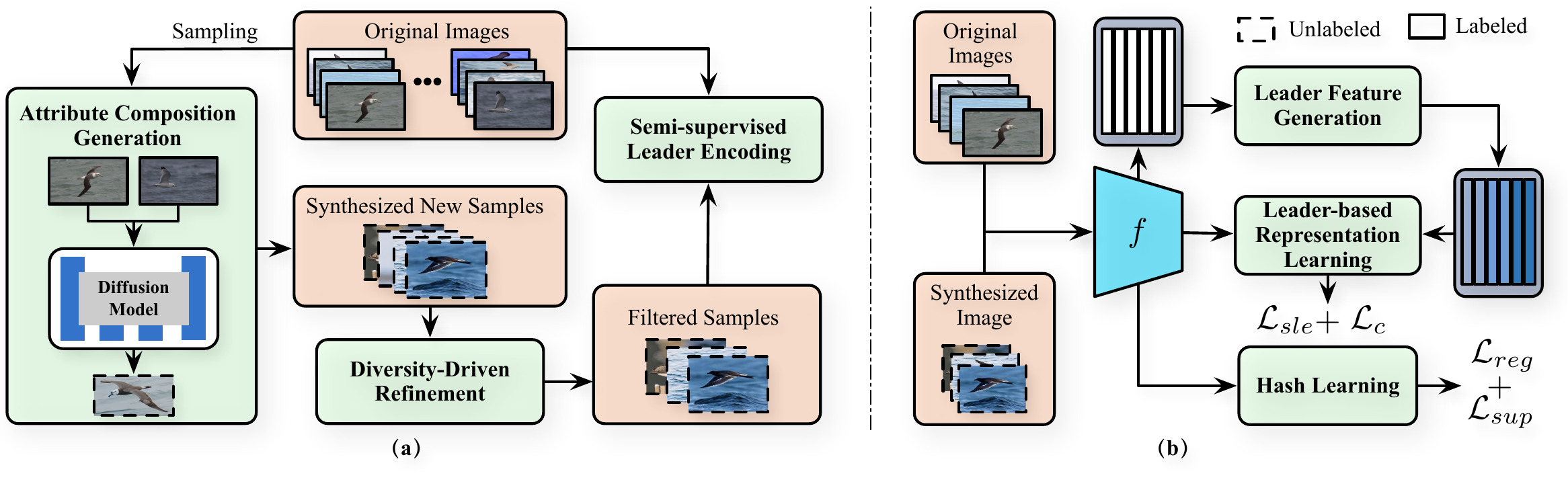}
  \vspace{-.30in}
  \caption{(a) Overview of our DiffGRE framework. First, we sample images from different categories to generate virtual-category images via our ACG. Then, we execute DDR to filter off the failure images and keep the remaining data for subsequential training. Last, we utilize SLE to exploit the generated data to further enhance the OCD models. (b) The pipeline of our SLE module. First, SLE mines the category-level relationship across the known and virtual categories for leader-based representation learning, following an iterative manner. After training, it leverages the generated leader feature as the known-category prior to performing robust online clustering inference. \label{fig:framework}}
  \vspace{-0.5em}
\end{figure*}
In this section, we first briefly introduce the OCD
task and then present our approach, DiffGRE, which integrates Generation, Refinement, and Encoding in a multi-stage fashion.

\par\noindent
\textbf{Problem Definition.}
The setting of OCD  is defined as follows. We are provided with a support set, denoted as $\mathcal{D}_S=\{(\mathbf{x}_i,y_i^s)\}_{i=1}^M\subseteq\mathcal{X}\times\mathcal{Y}_S$ for training, and a query set, denoted as $\mathcal{D}_Q=\{(\mathbf{x}_i,y_i^q)\}_{i=1}^N\subseteq\mathcal{X}\times\mathcal{Y}_Q$ for testing. Here, $M$ and $N$ represent the number of samples in $\mathcal{D}_S$ and $\mathcal{D}_Q$, respectively. $\mathcal{Y}_S$ and $\mathcal{Y}_Q$ are the label spaces for the support set and query set, respectively, where $\mathcal{Y}_S \subseteq \mathcal{Y}_Q$. We define classes in $\mathcal{Y}_S$ as known/old classes and classes in $\mathcal{Y}_Q / \mathcal{Y}_S$ as unknown/new classes. Only the support set $\mathcal{D}_S$ is used for model training. During testing, $\mathcal{D}_Q$ includes samples from both known and unknown categories, which are inferred one by one, allowing for online feedback.

\par\noindent
\textbf{Framework Overview.}
Here, we propose that DiffGRE synthesize new samples that include additional category knowledge to enhance the  OCD abilities of existing methods. \cref{fig:framework} shows an overview of DiffGRE’s framework. In particular, DiffGRE consists of three components: attribute composition generation, diversity-driven refinements, and semi-supervised leader encoding.

\enlargethispage{3mm}
\subsection{Attribute Composition Generation}
The Attribute Composition Generation (ACG) is proposed to synthesize virtual 
categories by interpolating the latent embeddings across different categories. The module is built based on previous Stable Diffusion Img2Img methods.
We apply the latent interpolation strategy that performs interpolation in both the CLIP's~\cite{radford2021learning} embedding space and the Stable Diffusion's latent space, facilitating attribute re-composition to synthesize novel images.

\par\noindent
\textbf{Stable Diffusion Img2Img.}
Different from vanilla diffusion models, the Stable Diffusion~\cite{rombach2022high} efficiently performs image generation and manipulation in the latent space, synthesizing images by progressively refining the latent representations. Specifically, given an image $I_i$, the encoder $\mathcal{E}$ maps it to a latent $z_i$, \emph{i.e.,} ${z_i} = \mathcal{E}(I_i)$.  
Besides, an additional input, \emph{e.g.,} a text prompt $t$ is fed to the model as a conditional input to control the synthesized image content. The text prompt is first pre-processed by a text tokenizer and then encoded by a text encoder, e.g., a pre-trained CLIP model.This can be formulated as: \emph{i.e.,} $z_{t} = \mathcal{T}(t)$ where $\mathcal{T}$ denotes the text encoder of the Stable Diffusion model.

\par\noindent
\textbf{Latent Interpolation in ACG.}
Based on the above settings, Stable Diffusion Img2Img variants can be applied to edit images with text prompts.
Previous works~\cite{ruiz2023dreambooth, gal2022image, wang2024enhance} in image editing have proposed using diffusion models to generate samples that have similar appearance attributes with known categories. However, this is incompatible with the OCD task. Our goal is to utilize the attributes of known categories to synthesize new samples belonging to virtual categories. To achieve this, we build on the Stable Diffusion Img2Img model by incorporating semantic latent codes from two inputs for attribute re-composition.
\par
In this work, we consider applying spherical interpolation~\cite{shoemake1985animating} in three embedding spaces, \ie, the latent space of Stable Diffusion, CLIP visual space, and CLIP textual space. Specifically, we use the VAE encoder in Stable Diffusion to extract the latent embeddings $\mathbf{z}^{l}$ and CLIP's visual encoder to extract visual embeddings $\mathbf{z}^{v}$. Meanwhile, we leverage a pre-trained image caption model~\cite{yu2022coca} to transform images into textural descriptions. Then, we extract textural embedding $\mathbf{z}^{t}$ from the textural descriptions via CLIP textual encoder. 
\par
The overall interpolation processes are formulated as:
\vspace{-.12in}

\begin{small}
\begin{equation}
\mathbf{\overline{z}^*} = \frac{\sin((1 - \lambda_{*}) \theta)}{\sin(\theta)} \mathbf{z}^{*}_{1} + \frac{\sin(\lambda_{*} \theta)}{\sin(\theta)} \mathbf{z}^*_{2}, ~ *\in \{ \textit{t}, \textit{v}, \textit{l} \} 
\label{si}
\end{equation}
\end{small}
\noindent where $\theta = \arccos(\mathbf{z}^*_{1} \cdot \mathbf{z}^*_{2})$
and \( \lambda_* \in [0, 1] \) is the interpolation parameter, and $\mathbf{z}^*_{1}, \mathbf{\bar{z}}^*_{2}$ denote the latent embeddings of two inputs, respectively. Following Stable Diffusion~\cite{rombach2022high}, we leverage both textual and latent embeddings to generate noise that is subsequently used for the denoising process. Meanwhile, inspired by~\cite{nichol2021glide}, we introduce CLIP visual embeddings to rectify the noise. The noise of the mixed latent is progressively removed at each timestep in the reverse denoising process, which is formulated as: 
\vspace{-.05in}
\begin{small}
\begin{equation}
\begin{gathered}
\mathcal{L}_{D M}=\mathbb{E}_{t, \mathbf{\overline{z}^l_0}, \epsilon}\left[\left\|\epsilon-\epsilon_\theta\left(\mathbf{\overline{z}^l}, t, \mathbf{\overline{z}^\textit{t}} \right)\right\|^2\right] \\
\mathbf{\overline{z}^l_t}=\sqrt{\alpha_t} \mathbf{\overline{z}^l_0}+\sqrt{1-\alpha_t} \epsilon, \epsilon \sim \mathcal{N}(0, \mathbf{I})
\end{gathered}
\end{equation}
\end{small}

\noindent where \( \epsilon_\theta \) is the UNet to predict noise, 
$\mathbf{\overline{z}^l_0}$ is the interpolated latent embedding of the input images, and $\alpha_t$ controls the noise schedule at timestep t.
The final latent embedding is passed to the VAE decoder to obtain the synthesized image.

\par\noindent\textbf{Discussion.} In the OCD task, transforming an image from a known category into an unknown category based on a text prompt is challenging because textual information for new categories is not directly available. As a result, directly applying text-to-image editing or text-to-image generative models to synthesize new samples for the OCD task is not a feasible approach. Meanwhile, the latent space in Stable Diffusion primarily captures fine-grained visual details for image reconstruction, whereas the CLIP latent space is optimized for high-level semantic understanding. Integrating CLIP embeddings enables Stable Diffusion to interpolate diverse attributes, blending characteristics from multiple categories to synthesize virtual categories.

\subsection{Diversity-Driven Refinement}
\label{sec:ddr}
Although ACG can generate numerous informative images, these images are not always favorable for OCD. As shown in \cref{tab:ablation1},
we experimentally find that direct training with all generated images disturbs discriminative feature extraction, especially for known categories. The primary reason may come from extensive and ambiguous synthesized samples that disrupt the original decision boundaries of known categories. To mitigate this issue, we propose the Diversity-Driven Refinement (DDR) module to retain samples that provide more additional information while filtering out those that do not, thereby reducing training overhead for the subsequent model training. Specifically, DDR removes samples that are too similar to known categories, preserving those that offer more diverse and valuable information.
\par
First, we leverage the training samples to measure the quality of generated images. One straightforward way is calculating the cosine similarity between generated samples and all training images. However, this approach becomes unstable when image features follow a long-tailed distribution. Hence, we chose to compute all similarity scores between generated samples and category centers that can reliably capture the feature distribution of the original training samples in the embedding space. Moreover, we empirically validate the effectiveness of the design in the Appendix. 
\par Given the similarities, we set a threshold, denoted as $\gamma$, to filter out the images with relatively high scores. The DDR can be divided into the following steps:
\begin{enumerate}
    \item Compute cosine similarity scores between generated images with original training category centers. The step is formulated as:
\vspace{-.05in}
\begin{small}
\begin{equation}
\text{s}(z_{\text{gen}}, c_k) = \frac{z_{\text{gen}} \cdot c_k}{\|z_{\text{gen}}\| \|c_k\|},
\end{equation}
\end{small}
\noindent where $z_{\text{gen}}$ represents the embedding of a generated image and $c_k$ represents the center for category $k$.

\item For each generated sample, calculate the mean score over all class centers. It can be represented as:
\vspace{-.05in}
\begin{small}
\begin{equation}
s_{mean}(z_{\text{gen}}) = \frac{1}{K} \sum_{k=1}^{K} \text{s}(z_{\text{gen}}, c_k), 
\end{equation}
\end{small}
where $K$ is the total number of category centers.

\item Remove all images based on the threshold.  Mathematically, it
can be formulated as:
\vspace{-.05in}
\begin{small}
\begin{equation}
    \mathcal{F}(I_{\text{gen}}) = \mathbf{1}( 
s_{mean}(z_{\text{gen}})
\leq \gamma) \cdot I_{\text{gen}},
\end{equation}
\end{small}
\noindent where $\mathcal{F}(I_{\text{gen}})$ is a function that decides whether to retain or remove the generated image $I_{\text{gen}}$.
\end{enumerate}

\subsection{Semi-Supervised Leader Encoding}
\par\noindent
\textbf{Motivation.} Although the combination of ACG and DDR is able to generate virtual category images, it is hard to assign precise category labels for each virtual image due to the randomness of attribute re-composition. We experimentally find that intuitively treating the generated images mixed by the same two categories as a unique new category leads to significant degradation of both known and unknown category discovery. Details are included in the Appendix. 

\par\noindent
\textbf{Virtual Category Assignment.} To fully mine the helpful knowledge from the generated virtual dataset, we propose a Semi-supervised Leader Encoding (SLE) approach to yield reliable category labels for virtual images. First, we combine the labeled data and generated data to build an agency training set $\mathcal{D}_{A} = \mathcal{D}_{S} \cup \mathcal{D}_{G}$. Secondly, we apply the off-the-shelf clustering method~\cite{rosvall2009map} on $\mathcal{D}_{A}$ and generate the virtual category labels $\mathcal{Y}_{A}$ for all data, resulting in $\mathcal{D}_{A}=\{(\mathbf{x}_i,y_i^s)\}_{i=1}^{M+ A}\subseteq\mathcal{X}\times\mathcal{Y}_A$, where $A$ is the number of synthesized image after DDR. Next, we leverage the ground-truth labels of partially-labeled data in $\mathcal{D}_{A}$ to rectify the clustering results further. Specifically, we first leverage the Hungarian optimal assignment algorithm~\cite{kuhn1955hungarian} to find the optimal mapping between the initial clustering assignment and the ground-truth labels of partially-labeled data. Based on the mapping results, we replace the clustering assignments of both labeled and unlabeled generated data according to the ground-truth labels and yield a new aligned version of the clustering assignments. Eventually, we establish a rectified $\mathcal{D}_{A}$ for OCD training.
\par\noindent
\textbf{Leader Feature Generation.} We calculate the average of the features that belong to the same category as a category-specific leader feature, $\mathbf{l}_{m}=\frac{1}{\left|\mathcal{Y}_{A}\right|} \sum_{\mathbf{x}_{m} \in \mathcal{Y}_{A}} f(\mathbf{x}_{m})$, where the $\left|\mathcal{Y}_{A}\right|$ is the number of virtual categories. Accordingly, the leader set for $\mathcal{D}_{A}$ is defined by: $\mathcal{C}_{A} = \{\mathbf{l}_{m}\}^{\left|\mathcal{Y}_{A}\right|}_{m=1}$. 

\par\noindent
\textbf{Leader-based Representation Learning.} To encourage models to learn robust features with a strong ability to discriminate different categories, we propose leader-based contrastive learning to increase the inter-category distance while reducing the intra-category distance on the virtual categories data. The objective is formulated as:
\vspace{-.05in}
\begin{small}
\begin{equation}
\mathcal{L}_{sle}\left(\mathbf{x}_{n}, y_{n}\right)=-\log \frac{\exp \left(f(\mathbf{x}_n) \cdot \mathbf{l}_{y_n}^{\mathrm{T}} / \tau\right)}{\sum_{m=1, m \neq y_n}^{|\mathcal{Y_{A}}|} \exp \left(f(\mathbf{x}_n) \cdot \mathbf{l}_{m}^{\mathrm{T}} / \tau\right)},
\end{equation}
\end{small}
\noindent where $\tau$ is the temperature factor and is empirically set to 0.05 for all experiments.
\par
Furthermore, inspired by the success of a combination of contrastive loss and classification loss in the other open-set task~\cite{zheng2019pyramidal}, we employ an additional classifier to supplement the leader-based contrastive learning with a cross-entropy loss, denoted as $\mathcal{L}_{c}$.
\par\noindent
\textbf{Remark.} Our SLE is tailor-made for the OCD task and brings three advantages: 1) SLE provides an effective manner to leverage the generated image with less noise, which injects additional category knowledge into OCD models, thereby resulting in better performance. 2) Unlike the existing OCD methods that apply representation learning on a low-dimensional hash space, our SLE reduces the information loss by performing contrastive learning on a high-dimensional feature space. 3) The calculated leader features can be stored as known-category prior, benefiting the subsequent on-the-fly inference in~\cref{sec:opt_inf}.

\subsection{Optimization and Inference}
\label{sec:opt_inf}

\noindent \textbf{Model Training.} 
We add our DiffGRE to the existing OCD method SMILE~\cite{ocd}. The SMILE method applies two hash heads to implement the OCD task. In order to enhance the representation embedding of old classes, we add another classifier to the final loss function. The total loss is:
\vspace{-.05in}
\begin{small}
\begin{equation}
\mathcal L = \mathcal L_{sup}+ \mathcal L_{reg} +\alpha * \mathcal L_{sle}+\beta*\mathcal L_{c},
\label{loss_total}
\end{equation}
\end{small}

\noindent where $L_{sup}$ and $L_{reg}$ are basic losses from SMILE~\cite{ocd}. Here, $L_{sup}$ is utilized for supervised contrastive learning, while $L_{reg}$ serves to constrain the output of the hash head $\mathcal{H}_h(\cdot)$ in~\cite{ocd} to approximate discrete values of +1 or -1. And $\alpha$ and $\beta$ weight the two losses to balance their contributions.

\par\noindent
\textbf{Discussion.} 
Existing OCD methods use hash-like mechanisms, which suffer from sensitivity, particularly for new categories. This is demonstrated by the results in~\cref{tab:hash_based}, where the hash-based methods are shown to be more sensitive compared to PHE~\cite{zheng2025prototypical}. To address this, we design Online Cluster Inference (OCI) with high-dimensional features, which preserves rich information while fully leveraging the learned leader representation in SLE and mitigating the sensitivity issues.

\par\noindent
\textbf{Online Cluster Inference based on SLE.}
In the testing stage, we first establish a dynamic leader memory, which is initialized by the category-specific leader features in~\cref{sec:ddr}. Then, we calculate the maximal intra-category distance, $\Delta_{max}$, as an adaptive threshold to determine if this instance belongs to unknown categories following an online clustering fashion. Given a test instance, $\mathbf{x}_{j} \in \mathcal{D}_{Q}$, if it is predicted to belong to known categories, we assign its estimated category label $\hat{y}_{j}$ to the category label according to the nearest leader. Otherwise, we create a new leader with the instance feature and append it to the dynamic leader memory. During the testing process, we update both known- and unknown-category leaders by momentum averaging of corresponding instance features. The algorithm is described in the Appendix.

\section{Experiments}
\label{sec:experiment}

\subsection{Experiment Setup}
\noindent \textbf{Datasets.} We have conducted experiments on six fine-grained datasets, including CUB-200~\cite{cub}, Stanford Cars~\cite{scars}, Oxford-IIIT Pet~\cite{pets}, and three super-categories from the more challenging dataset, iNaturalist~\cite{inaturalist}, including Arachnida, Animalia, and Mollusca. Following the setup in OCD~\cite{ocd}, the categories of each dataset are split into subsets of known and unknown categories. Specifically, 50\% of the samples from the seen categories are used to form the support set, $\mathcal{D}_S$ for training, while the remainder forms the query set $\mathcal{D}_Q$ for on-the-fly testing. Details about the datasets are provided in the Appendix.

\noindent \textbf{Evaluation Metrics.} We follow~\cite{ocd} and adopt clustering accuracy as the evaluation metric, $ACC = \frac{1}{|\mathcal{D}_Q|} \sum_{i=1}^{|\mathcal{D}_Q|} \mathbb{I}(y_i = C(\hat{y}_i))$, where $\hat{y}_i$ represents the predicted labels and $y_i$ denotes the ground truth. The function $C$ denotes the optimal permutation that aligns predicted cluster assignments with the actual class labels. We report ``ACC-ALL'', ``ACC-OLD'' and ``ACC-NEW'' for all categories, known categories, and unknown categories.

\noindent \textbf{Implementation Details.} For a fair comparison, we follow OCD~\cite{ocd} and use the DINO-pretrained ViT-B-16~\cite{dino} as the backbone. During training, only the final block of ViT-B-16 is fine-tuned. The Projector $\mathcal{H}_h(\cdot)$ consists of three linear layers with an output dimension set to $L=12$, which produces $2^{12} = 4096$ binary category encodings. We follow OCD to set this dimension for a fair comparison. 
The ratio $\alpha$ and $\beta$ in the total loss are set to 0.3 and 1.0 for all datasets. 
More details about $\gamma$ analysis are provided in ~\cref{fig:hyperparameters1}.

\par\noindent
\textbf{Inference Strategy.} Existing OCD methods mainly employ the \textit{hash-like inference} strategy, in which the image features are mapped to hash-like category descriptors. Then, they group images with the same descriptors into the same class, thereby implementing the on-the-fly inference. However, this strategy inevitably loses accuracy due to binarization processing and limited hash length. In contrast, in this paper, we introduce the Online Clustering Inference (OCI) strategy based on our SLE. We continuously measure the similarities between stream inputs and category-specific leader features, dynamically assigning new inputs to known categories or creating new categories for them.

\begin{table*}[!t]
  \caption{Comparison with \textbf{hash-like inference} methods, with the best results in \textbf{bold} and the second best \underline{underlined}.}
  \setlength{\tabcolsep}{7.5pt}
  \renewcommand{\arraystretch}{1}
  \label{tab:hash_based}
  \centering
  \footnotesize 
  \begin{tabular}{l|ccc|ccc|ccc|ccc} 
    \toprule
    \multirow{2}{*}{Method} & \multicolumn{3}{c|}{\textbf{Arachnida}} & \multicolumn{3}{c|}{\textbf{Mollusca}} & \multicolumn{3}{c|}{\textbf{Oxford Pets}}  & \multicolumn{3}{c}{\textbf{Average}} \\
    \cline{2-13}
    & All & Old & New & All & Old & New & All & Old & New & All & Old & New \\
    \midrule

BaseHash~\cite{ocd} &  23.6 
 & 42.3
 &  10.8
& 25.7  &   40.0 &  18.0& 35.8  &  36.1  & 35.7 & 28.4 &  39.5 &  21.5 \\

BaseHash~\cite{ocd} + DiffGRE (Ours) & 27.2 & 50.5
 & 12.4 & 28.8& 40.6 
 & 20.9
 & 38.6 & 39.2 & 39.2 & 31.5    & 43.4  &    24.2  \\
\cline{1-13}
SMILE~\cite{ocd} & 27.9
 & 53.0 &  12.2 & 33.5 & 42.8 & \underline{28.6} & 41.2 & 42.1 & 40.7 &  34.2   & 46.0   & 27.2 \\
SMILE~\cite{ocd} + DiffGRE (Ours) & 35.4 & 66.8 & \textbf{15.6} & 36.5 & 44.2& \textbf{32.5} & 42.4 & 42.1 & 42.5 & 38.2  & 51.0    & \textbf{30.2} \\
\cline{1-13}
PHE~\cite{zheng2025prototypical} & \underline{37.0} & \underline{75.7} & \underline{12.6} & \underline{39.9} & \textbf{65.0} &26.5 & \underline{48.3} & \textbf{53.8} & \underline{45.4} & \underline{41.7}& \textbf{64.8} &  28.2\\
PHE~\cite{zheng2025prototypical} + DiffGRE (Ours) &  \textbf{39.1} & \textbf{76.5} & \textbf{15.6}  &  \textbf{40.3} & \underline{63.4} & 28.1  &  \textbf{48.6} & \underline{52.6} & \textbf{46.6}  & \textbf{42.7} & \underline{64.2} & \underline{30.1}   \\
    \bottomrule
    \multirow{2}{*}{Method} & \multicolumn{3}{c|}{\textbf{CUB}} & \multicolumn{3}{c|}{\textbf{Stanford Cars}} & \multicolumn{3}{c|}{\textbf{Animalia}} & \multicolumn{3}{c}{\textbf{Average}} \\
    \cline{2-13}
    & All & Old & New & All & Old & New & All & Old & New & All & Old & New \\
    \midrule

BaseHash~\cite{ocd} & 21.1 & 26.2 & 18.5 & 15.4 &23.0 &11.7&  31.2 & 49.7& 23.6  & 22.6  & 33.0 &    17.9 \\

BaseHash~\cite{ocd} + DiffGRE (Ours) &  34.2 &  48.8  & 26.9 & 26.9 & 47.3  & \textbf{17.0} & 36.4 & 60.2  & 26.6 & 32.5  &  52.1   & 23.5\\

\cline{1-13}
SMILE~\cite{ocd} & 32.2 & 50.9 & 22.9 &  26.2 & 46.7& 16.3 & 34.8 & 58.8 & 24.9 & 31.1   &  52.1 &   21.4 \\

SMILE~\cite{ocd} + DiffGRE (Ours) & 35.4 & \textbf{58.2} & 23.8 & 30.5 & 59.3 & 16.5 & 37.4 & \textbf{69.3} & 24.3 & 34.4  & \textbf{62.3} &   21.5 \\
\cline{1-13}
PHE~\cite{zheng2025prototypical} & \underline{36.4} &55.8 &\underline{27.0}& \underline{31.3}& \underline{61.9}& 16.8 & \underline{40.3} & 55.7 &\underline{31.8} & \underline{36.0} &57.8 &\underline{25.2}\\
PHE~\cite{zheng2025prototypical}+ DiffGRE (Ours) & \textbf{37.9} & \underline{57.0} & \textbf{28.3} &  \textbf{32.1} &\textbf{63.3} &\underline{16.9}  & \textbf{42.0}  & \underline{64.5}  & \textbf{32.6} & \textbf{37.3} &  \underline{61.6} & \textbf{25.9}   \\

    \bottomrule
  \end{tabular}
  \vspace{-0.5em}
\end{table*}

\begin{table*}[!t]
  \caption{ Comparison with \textbf{online clustering inference} methods, with the best results in \textbf{bold} and the second best \underline{underlined}.}
  \setlength{\tabcolsep}{7.5pt}
  \renewcommand{\arraystretch}{1}
  \label{tab:feature_based}
  \centering
  \footnotesize 
  \begin{tabular}{l|ccc|ccc|ccc|ccc} 
    \toprule
    \multirow{2}{*}{Method} & \multicolumn{3}{c|}{\textbf{Arachnida}} & \multicolumn{3}{c|}{\textbf{Mollusca}} & \multicolumn{3}{c|}{\textbf{Oxford Pets}}  & \multicolumn{3}{c}{\textbf{Average}} \\
    \cline{2-13}
    & All & Old & New & All & Old & New & All & Old & New & All & Old & New \\

    \midrule

BaseHash~\cite{ocd} + SLC~\cite{slc} &25.4 & 44.6 & 11.4 &31.1 & 59.8 & 15.0  & 35.5 &41.3 & 33.1 & 30.7   & 48.6 &  19.8  \\

BaseHash~\cite{ocd} + SLE-based (Ours) & 42.0  &   55.0  &  33.9
 & 40.7  & 49.6 & 36.0 &   45.5 &  48.0 & 44.0&  42.7  & 50.9 & 38.0 \\
 BaseHash~\cite{ocd} + DiffGRE (Ours) & 44.9 
 &60.7 
  &  34.9 & 41.0 &  49.0 
 & \underline{36.7} & 46.3 & 43.0 & 48.1
 & 44.1 & 50.9 & 39.9 \\

\cline{1-13}

SMILE~\cite{ocd} + SLE-based (Ours) & 43.7 & 60.0 & \underline{37.0}
 &   \underline{42.7} 
   &  54.3
 &  36.6  &   49.1  &  49.1   & 49.1
 &   45.2   &54.5     & \underline{40.9} \\
SMILE~\cite{ocd} + DiffGRE (Ours) & \underline{45.3} & 56.6 & \textbf{40.6} & \textbf{43.6} &  51.0  & \textbf{39.6} & \textbf{51.1} & \textbf{54.7} & \underline{49.2} & \textbf{46.7}  & 54.1 &   \textbf{43.1}
\\
\cline{1-13}
PHE~\cite{zheng2025prototypical} + SLE-based&  44.9 & \textbf{78.9} & 23.5 &  40.3 & \underline{60.8} & 29.5   & 46.2 & 44.4 & 47.1 & 43.8 & \underline{61.4} & 33.4   \\
PHE~\cite{zheng2025prototypical} + DiffGRE (Ours) &  \textbf{47.7} & \underline{76.6} & 29.4 & 42.6 & \textbf{62.0} & 32.3  & \underline{49.6} & \underline{50.1} &\textbf{49.3} &  \underline{46.6} & \textbf{62.9} & 37.0   \\

    \bottomrule
    \multirow{2}{*}{Method} & \multicolumn{3}{c|}{\textbf{CUB}} & \multicolumn{3}{c|}{\textbf{Stanford Cars}} & \multicolumn{3}{c|}{\textbf{Animalia}} & \multicolumn{3}{c}{\textbf{Average}} \\
    \cline{2-13}
    & All & Old & New & All & Old & New & All & Old & New & All & Old & New \\
    \midrule

BaseHash~\cite{ocd} + SLC~\cite{slc} & 28.6 & 44.0  &20.9 &14.0 &23.0 &9.7 & 32.4 & \underline{61.9} &19.3 &   25.0  
 &  43.0    & 16.6
   \\

BaseHash~\cite{ocd} + SLE-based (Ours) &  37.3  &  38.4   &   36.7  & 21.7
 &  30.3 
&  17.5
&  44.9  &  57.3  &  39.8
 &  34.6    & 42.0   &31.3  \\
BaseHash~\cite{ocd} + DiffGRE (Ours) & 38.5 & 40.6  & \underline{37.5} & 26.0 & 37.9 
 & \textbf{20.3} & 47.4  & 59.6  &  42.3 & 37.3   &    46.0    & \underline{33.4} \\

\cline{1-13}

SMILE~\cite{ocd} + SLE-based  (Ours)  & 39.3 &  43.4 & 37.2 & 27.5   & 47.4 & 18.0
 &   \underline{48.8} 
 & 61.0 
 & \underline{43.8}
 & \underline{38.5}     & 50.6   &   33.0  \\
SMILE~\cite{ocd} + DiffGRE (Ours) & \textbf{42.8} & 47.8 & \textbf{40.2} & \textbf{28.3} & \textbf{49.3}  & \underline{18.1} & \textbf{49.2} & 61.1 & \textbf{44.3} & \textbf{40.1}    & \textbf{52.7}   & \textbf{34.2} \\
\cline{1-13}
PHE~\cite{zheng2025prototypical}+ SLE-based  (Ours)  & 41.0 & \underline{52.1} & 35.5 &  25.2 & 44.5 & 15.9 &  40.5 & 61.6 & 32.9 &   35.6 & \textbf{52.7}  & 28.1  \\
PHE~\cite{zheng2025prototypical} + DiffGRE (Ours) & \underline{42.5} & \textbf{54.4} & 36.5 &  \underline{27.7} & \underline{48.1} & 17.8   &  43.5 & \textbf{63.2} & 35.3  &  37.9 & \underline{55.2} & 29.9    \\

    \bottomrule
  \end{tabular}
  \vspace{-.1in}
\end{table*}

\subsection{Comparison with State-of-the-Art Methods}

As the proposed SLE module enables hash-like inference methods to address OCD issues through online clustering inference (OCI), we denote testing SMILE with SLE-based OCI strategy as ``SMILE + SLE-based''. Meanwhile, ``SMILE + DiffGRE'' can achieve on-the-fly inference in both hash-like and OCI strategies. For a thorough and fair comparison, we conducted detailed experiments under these two inference strategies to verify our plug-and-play DiffGRE framework.

\par\noindent
\textbf{Results of Hash-like Inference}. The results compared with hash-like inference methods are shown in ~\cref{tab:hash_based}, where we can clearly observe that the introduction of the DiffGRE framework significantly enhances the performance of these strong baselines across all metrics. For example, the addition of DiffGRE brings an average improvement of 6.5\% in ACC-ALL and 11.5\% in ACC-OLD for BaseHash across six datasets. Similarly, SMILE can benefit from our DiffGRE with average gains of 3.7\% in ACC-ALL and 7.6\% in ACC-OLD across six datasets. These promising results demonstrate that our DiffGRE, as a plug-and-play module, effectively elevates the on-the-fly discovery accuracy of existing hash-like inference methods. Moreover, the combination of ``PHE + DiffGRE" consistently outperforms all competitors, achieving the highest ACC-ALL averages.

\par\noindent
\textbf{Results of Online Clustering Inference}. The results compared with Online Clustering Inference methods are shown in \cref{tab:feature_based}. First, our SLE enables hash-like inference methods to perform online clustering inference, and the online clustering inference based on our SLE module achieves higher accuracy than the leading competitor, SLC. For example, ``BaseHash + SLE-based'' exceeds ``BaseHash + SLC'' by an average of 10.8\% in ACC-ALL and 16.5\% in ACC-New across six datasets. Moreover, the incorporation of our full DiffGRE framework consistently enhances all methods. Specifically, it boosts ``BaseHash + SLE-based'' by an average of 2.1\% in ACC-ALL, and enhances ``SMILE + SLE-based'' by an average of 1.6\% in ACC-ALL. Notably, although our DiffGRE improves PHE in ``PHE + DiffGRE'' setting, the improvement is relatively lower than that under the hash-like inference strategy. This is because PHE focuses on designing a superior hash network, where the feature extractor could not be appropriately optimized. As a consequence, the improvements are limited by unsatisfactory features used for inference in OCI strategy.

\par\noindent
\enlargethispage{3mm}
\textbf{Comprehensive evaluation.} Our DiffGRE achieves consistent improvements in all datasets in both hash-like and OCI inference strategies. Especially with the OCI strategy, our method achieves a new state-of-the-art performance on the average accuracy over six datasets. These results verify the discussion in~\cref{sec:opt_inf}.

\subsection{Ablation Study}

We report an ablation analysis of our DiffGRE on the Arachnida, Mollusca and CUB datasets, as shown in \cref{tab:ablation1}. For evaluation, we use hash-like inference. ``SMILE + DiffGRE" refers to the complete method achieved by applying our plug-and-play DiffGRE framework to SMILE. The variant ``$w/o$ SLE" is obtained by removing the SLE module, resulting in an average reduction of 3.6\% in ACC-ALL across the three datasets. 
This confirms that the SLE module effectively introduces additional category knowledge into the model training process. Removing the DDR module leads to an average decrease of 2.3\% in ACC-ALL and 4.6\% in ACC-OLD across the datasets, indicating that low-quality samples synthesized by diffusion models can disrupt discriminative feature extraction. This also underscores the effectiveness of the DDR module in filtering out low-quality samples. Additionally, compared to SMILE, the ``$w/o$ DDR" variant shows an average improvement of 3.2\% across the three datasets, demonstrating that introducing additional category information, even if it includes low-quality samples, helps in identifying fine-grained categories and enhancing category discovery accuracy.

\begin{table}[!t]
  \setlength{\tabcolsep}{2pt}
  \footnotesize
  \renewcommand{\arraystretch}{0.8}
    \centering
    \caption{Ablation study on training components.\label{tab:ablation1}}
    \begin{tabular}{l|ccccccccc}
      \toprule
      \multirow{2}{*}{Method}  & \multicolumn{3}{c}{Arachnida} & \multicolumn{3}{c}{Mollusca} & \multicolumn{3}{c}{CUB}\\
      \cmidrule(lr){2-4} \cmidrule(lr){5-7}
      \cmidrule(lr){8-10}
       & All & Old & New & All & Old & New & All & Old & New \\
      \midrule
   SMILE~\cite{ocd}      &  27.9 & 53.0 & 12.2& 33.5 & 42.8 & 28.6 & 32.2 &50.9 & 22.9 \\
$w/o$ $\mathcal L_{sle}$ &  29.3 & 53.7 & 13.9 & 33.9  & 41.4 &  29.9 & 33.2 & 53.9 &22.9
\\
  $w/o$ $\mathcal L_{c}$   &  34.5  &  66.3 &   14.6 & 36.0 & 42.9 & 32.3 & 33.6 & 55.4 & 22.7\\
  $w/o$ DDR    & 33.5 & 62.8 & 15.2 & 34.5& 39.4 & 31.9 &  32.4&  53,2 & 22.0 \\
   SMILE + DiffGRE&\textbf{35.4}&\textbf{66.8}&\textbf{15.6}&\textbf{36.5}&\textbf{44.2}&\textbf{32.5 } & \textbf{35.4} & \textbf{58.2} & \textbf{23.8}\\
      \bottomrule
    \end{tabular}
  \vspace{-2em}
\end{table}

\begin{figure*}[!t]
  \centering
  \includegraphics[width=\linewidth]{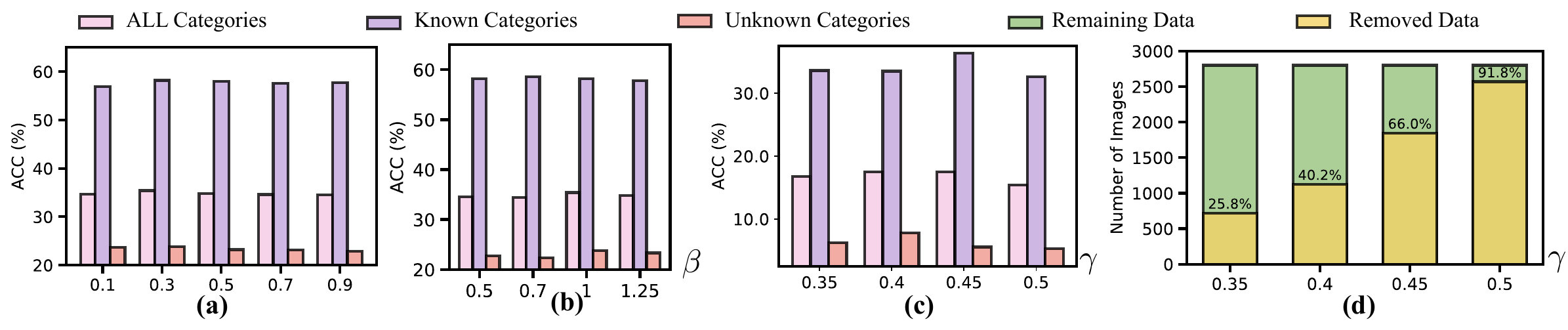}
  \vspace{-1em}
  \caption{Impact of hyperparameters, \textit{i.e.,} $\alpha$, $\beta$ and $\gamma$. We implement experiments on CUB~\cite{cub} in \textbf{(a-b)} and Arachnida~\cite{inaturalist} in \textbf{(c-d)}. 
  \label{fig:hyperparameters1}}
  \vspace{-.05in}
\end{figure*}

\begin{figure*}[!t]
  \centering
  \includegraphics[width=\linewidth]{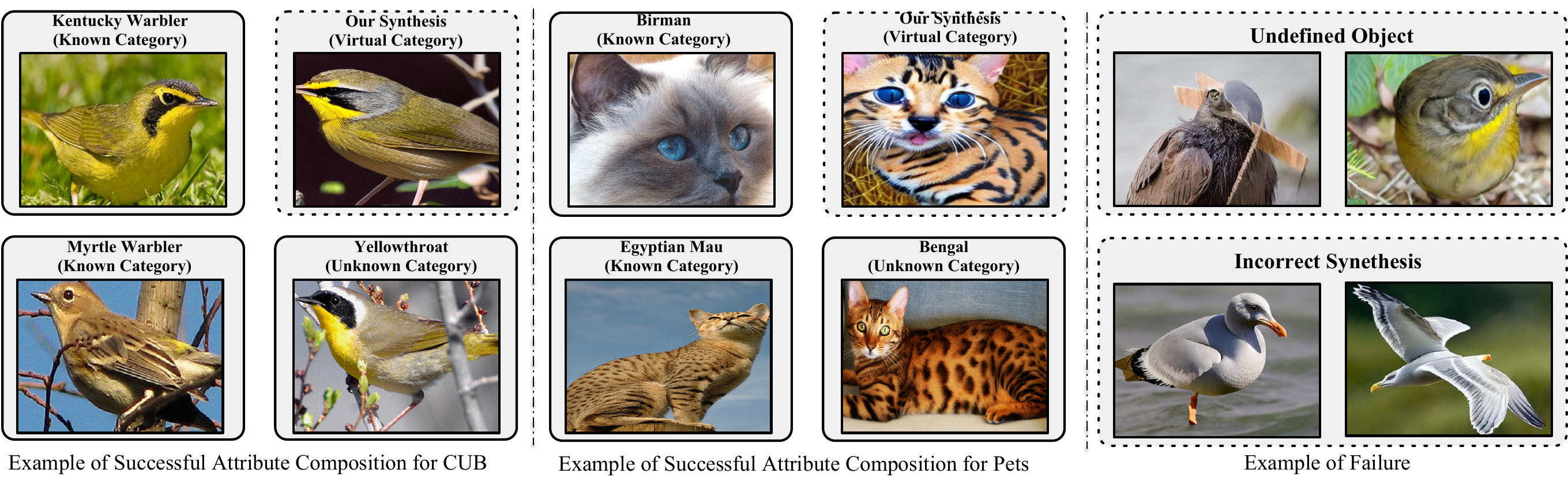}

  \caption{Visualization of the synthesized images. We illustrate two examples of successful attribute composition on the CUB~\cite{cub} and Pets~\cite{pets} datasets. Each successful example includes a synthesized sample and a real sample belonging to an unknown category, which is extracted through the nearest neighbor search. The nearest neighbor search is performed on both the known and unknown category images with respect to the synthesized samples. We show examples of failed generations: \textit{undefined object} and \textit{incorrect synthesis}.\label{fig:visualization}}
  \vspace{-1em}
\end{figure*}

\subsection{Analysis}

\noindent\textbf{Hyper-Parameters Analysis.} 
1) The impact of the ratios $\alpha$ and $\beta$ in ~\cref{loss_total} is illustrated in \cref{fig:hyperparameters1} (a-b). Parameters $\alpha$ and $\beta$ control the relative importance of $L_{sle}$ and $L_c$ during the training process, respectively. The parameter $\alpha$ is relatively insensitive, showing subtle variation in accuracy when adjusted from 0.1 to 0.9. We fine-tuned this parameter on the CUB dataset and fixed it at 0.3 for all datasets. Similarly, $\beta$ remains stable across variations, as shown in ~\cref{fig:hyperparameters1} (b). We set $\beta = 1.0$ for all datasets during training.
2) We examine the impact of $\gamma$ in the DDR module on the Arachnida dataset. This parameter controls the filtering of synthesized samples, where a lower $\gamma$ results in fewer samples being filtered. As shown in ~\cref{fig:hyperparameters1} (c) and (d), when the number of the remaining images is comparable to that of labeled data, the model reaches the best accuracy on unknown categories. Thus, we set 0.4 for the Arachnida dataset. Similarly, we set $\gamma$ for other datasets according to their number of labeled data. Moreover, we explore the impact of interpolation parameters $ \lambda_*$ in~\cref{si}, where $*\in \{ \textit{t}, \textit{v}, \textit{l} \}$. We follow the previous work~\cite{ramesh2022hierarchical} to set $\lambda_t=0.7$. Then, we fix $\lambda_t$ to tune the optimal $\lambda_v=0.7$ and $\lambda_l=0.8$ (see detailed analysis in the Appendix).

\par\noindent
\textbf{Impact of Different Synthesis Methods.}
To further validate the effectiveness of our method, we compare it with two other approaches for synthesizing new samples. The first includes T2I generative models, Diff-Mix~\cite{wang2024enhance} and Da-Fusion~\cite{trabucco2023effective}, using known categories as text inputs. For Diff-Mix, we also use unknown categories as text inputs, but note that this is not allowed in the OCD task. The second approach involves pixel-level mix methods, CutMix~\cite{yun2019cutmix} and MixUp~\cite{zhang2018mixup}, which, together with the T2I models, are used to replace our ACG for a fair comparison.

The results are summarized in \cref{tab:generative_models}. As shown, pixel-level methods perform worse than our approach. While T2I models generate high-quality images, their improvement for unknown categories is limited, proving that Diff-Mix can only synthesize images within known categories, making it inefficient for the OCD task. In contrast, our method achieves higher accuracy, demonstrating its effectiveness in synthesizing novel samples for unknown categories. Additionally, we compare the feature spaces generated by our method and Diff-Mix, with detailed results in the Appendix.

\begin{table}[t]
 \setlength{\tabcolsep}{3pt}
\renewcommand{\arraystretch}{0.8}
\footnotesize
\centering

\caption{Impact of different synthesis methods.}

  \label{tab:generative_models}

  \begin{tabular}{l|ccc} 
    \toprule
    \multirow{2}{*}{Method} & \multicolumn{3}{c}{\textbf{CUB}} \\
    \cline{2-4}
    & All & Old & New \\
    \midrule
    SMILE~\cite{ocd} & 32.2 & 50.9 & 22.9 \\
    SMILE~\cite{ocd} + Da-Fusion~\cite{trabucco2023effective} & 34.4  & 56.7  & 23.2 \\
    SMILE~\cite{ocd} + Diff-Mix~\cite{wang2024enhance}
    (w/ known category)  & 32.9 & 54.4 &22.1\\
    SMILE~\cite{ocd} + Diff-Mix~\cite{wang2024enhance} (w/ unknown category)  & 35.1 & 56.6  & 24.3 \\
    SMILE~\cite{ocd} + CutMix~\cite{yun2019cutmix} &32.8 & 54.6 & 21.9 \\
    SMILE~\cite{ocd} + MixUp~\cite{zhang2018mixup} & 33.3 & 55.6 & 22.1\\
    SMILE~\cite{ocd} + DiffGRE (Ours) & 35.4 & 58.2 & 23.8 \\
    \bottomrule
  \end{tabular}
  \vspace{-0.5em}

\end{table}

\subsection{Visualization and Limitation}

\textbf{Visualization}. We also visualize the generated images on different datasets in~\cref{fig:visualization}. Subjectively, the images generated by our ACG show clear attribute features and preserve the main characteristic of the object. Interestingly, the nearest neighbor of the generated image often belongs to the unknown categories in the OCD benchmark. Several failure examples are provided. We categorize generated results with noticeable errors or defects as ``incorrect'' and those that are entirely unrecognizable as ``undefined''. These failure cases are mainly caused by the randomness of the diffusion models (see more examples and comparisons in the Appendix). 

\section{Conclusion}
\label{sec:conclusion}
This work focuses on a practical and challenging OCD task. To improve the existing OCD models that lack sufficient category knowledge, we propose to embrace pre-trained generative models based on diffusion models and accordingly design a new plug-and-play DiffGRE framework to synthesize new training samples. DiffGRE aims to generate new samples that include additional category knowledge and then effectively leverage these images to enhance OCD models. Extensive experiments validate the effectiveness of our methods over six datasets.
{
    \small
    \bibliographystyle{ieeenat_fullname}
    \bibliography{main}
}


\definecolor{darkorange}{rgb}{1.0, 0.55, 0.0}
\newcommand{\nan}[1]{{\color{darkorange}#1}}
\definecolor{darkbluee}{rgb}{0.69, 0, 0.11}
\newcommand{\zhy}[1]{{\color{darkbluee}#1}}
\definecolor{remark}{rgb}{1,.5,0} 
\definecolor{citecolor}{rgb}{0,0.443,0.737} 
\definecolor{linkcolor}{rgb}{0.956,0.298,0.235} 

\newpage
\maketitlesupplementary

\setcounter{page}{1}

\renewcommand\thesection{\Alph{section}}

\setcounter{section}{0}
\setcounter{figure}{0}
\setcounter{table}{0}

\numberwithin{equation}{section}
\numberwithin{figure}{section}
\numberwithin{table}{section}

\etocdepthtag.toc{mtappendix}
\etocsettagdepth{mtchapter}{none}
\etocsettagdepth{mtappendix}{subsection}

\renewcommand{\contentsname}{Appendix}
\tableofcontents

\section{Implementation Details}

\subsection{Datasets Details}
\label{sub_sec:datasets}
\noindent \textbf{Dataset Details.}
As shown in Table~\ref{sub_tab:dataset}, we evaluate our method across multiple benchmarks, including the introduction of the iNaturalist 2017~\cite{inaturalist} dataset to the On-the-Fly Category Discovery (OCD) task. This demonstrates the robustness of our approach in addressing challenging fine-grained datasets. The iNaturalist 2017 dataset, collected from the citizen science platform iNaturalist, comprises 675,170 training and validation images spanning 5,089 fine-grained natural categories, including Plantae (plants), Insecta (insects), Aves (birds), and Mammalia (mammals), distributed across 13 super-categories. These super-categories exhibit substantial intra-category variation, posing significant challenges for fine-grained classification.
\subsection{Evaluation Metric Details}
For our evaluation, we focus on three super-categories: Arachnida, Animalia, and Mollusca. Following the OCD protocol~\cite{ocd}, the categories within each dataset are split into seen and unseen subsets. Specifically, 50\% of the samples from seen categories form the labeled training set $\mathcal{D}_S$, while the remaining samples are included in the unlabeled set $\mathcal{D}_Q$ for on-the-fly testing.

\begin{table}[t]
  \setlength{\tabcolsep}{4pt}
  \renewcommand{\arraystretch}{1}
  \caption{Statistics of datasets used in our experiments.}
  \label{sub_tab:dataset}
  \centering
\resizebox{0.47\textwidth}{!}{
  \begin{tabular}{l|cccccc}
    \toprule
    & CUB & Scars & Pets  & Arachnida & Animalia & Mollusca \\
    \midrule
    $|Y_S|$ & 100 & 98 & 19  & 28 & 39  & 47  \\
    $|Y_Q|$ & 200 & 196 & 38  & 56 & 77 & 93\\
    \midrule
    $|\mathcal D_S|$ & 1.5K & 2.0K & 0.9K  & 1.7K & 1.5K & 2.4K \\
    $|\mathcal D_Q|$ & 4.5K & 6.1K & 2.7K  & 4.3K & 5.1K & 7.0K \\
    \bottomrule
  \end{tabular}

} 
\end{table}
\subsection{Algorithm Pipeline for SLE-based Inference}

In the testing stage, we first establish a dynamic leader memory, which is initialized by the category-specific leader features in Sec. 3.2.

Then, we calculate the maximal intra-category distance, $\Delta_{max}$, as an adaptive threshold to determine if this instance belongs to unknown categories following an online clustering fashion. Given a test instance, $\mathbf{x}_{j} \in \mathcal{D}_{Q}$, if it is predicted to belong to known categories, we assign its estimated category label $\hat{y}_{j}$ to the category label according to the nearest leader. Otherwise, we create a new leader with the instance feature and append it to the dynamic leader memory. During the testing process, we update both known- and unknown-category leaders by momentum averaging of corresponding instance features. The algorithm is deliberated in Algorithm~\ref{alg:inference}.
\begin{algorithm}[t]
\small
\caption{On-the-Fly Inference based on SLE.\label{alg:inference}}
\KwIn{Test data $\mathcal{D}_Q$, trained backbone network $f(\cdot)$, maximal intra-category distance $\Delta_{max}$, and a set of known-category leader features $\mathcal{C}_{S} = \{\mathbf{l}_{i}\}_{i=1}^{\left| \mathcal{Y}_{S}\right|}$}

\For{ $\mathbf x_i \in \mathcal{D}_Q$}{ 
        Extract instance feature  $f(\mathbf{x}_{i})$\;
        \For{leader feature  $\mathbf{l}_{j} \in \mathcal{C}_{S}$}{
            Compute $l_{2}$ distance $||\mathbf{l}_{j} - f(\mathbf{x}_{i})||_{2}^{2}$ \;
            \If{$ ||\mathbf{l}_{j} - f(\mathbf{x}_{i})||_{2}^{2} \geq \Delta_{max}$}{
            Add $f(\mathbf{x}_{i})$ to $\mathcal{C}_{S}$ \;
            \textbf{Return} $\hat{y}_{j}=|\mathcal{C}_{S}| + 1$
             \tcp*{Create a new category}
            }  
        }
        $\hat y_i = argmin_{j} ||\mathbf{l}_{j} - f(\mathbf{x}_{i})||_{2}^{2}$ \;
        $\mathcal{C}\left[\hat{y}_{i}\right]=\eta \cdot \mathcal{C} \left[\hat{y}_{n}\right]+(1-\eta) \cdot f(\mathbf{x}_{i})$ \;
        \textbf{Return} $\hat y_i$
        \tcp*{belong to a known category}
}
\end{algorithm}

\subsection{Details of the Compared Methods.}
\label{sup_sec:methods}
Since the OCD task requires real-time inference and is relatively new, traditional baselines from NCD and GCD are inappropriate for this scenario. Thus, we selected the SMILE model~\cite{ocd} as a comparative baseline and included three hash-like competitive methods, SMILE baseline~\textbf{(BaseHash)}~\cite{ocd}, Prototypical Hash Encoding ~\textbf{(PHE)}~\cite{zheng2025prototypical}, Ranking Statistics~\textbf{(RankStat)}~\cite{rs}  and Winner-take-all~\textbf{(WTA)}~\cite{wta}, and one online clustering method ~\textbf{Sequential Leader Clustering (SLC)}~\cite{slc} for evaluation. The elaborated introductions are follows:
\begin{itemize}
    \item \textbf{Sequential Leader Clustering (SLC)}~\cite{slc}: This traditional clustering method is tailored for sequential data analysis.
    \item \textbf{Ranking Statistics (RankStat)}~\cite{rs}: RankStat identifies the top-3 indices in feature embeddings to serve as category descriptors.
    \item \textbf{Winner-take-all (WTA)}~\cite{wta}: WTA utilizes the indices of the highest values within groups of features as the basis for category description. These three robust baselines are established in accordance with the SMILE.
    \item 
    \textbf{Prototypical Hash Encoding}~\textbf{(PHE)}~\cite{zheng2025prototypical}: PHE leverages Category-aware Prototype Generation (CPG) to capture intra-category diversity and Discriminative Category Encoding (DCE) to enhance hash code discrimination, ensuring effective category discovery in streaming data.
\end{itemize}

\subsection{SMILE Approach and Training Details}

We use the pre-trained Stable Diffusion v1.4 to fuse semantic latent embeddings without any fine-tuning. The OCD models (\ie, based projector $\mathcal{H}(\cdot)$, hash projector $\mathcal{H}_h(\cdot)$~\cite{ocd} and ViT-B-16~\cite{dino} backbone) are optimized by SGD~\cite{qian1999momentum} optimizer during training process. We set the initial learning rate to be $1e-2$ with a batch size of 128. A weight decay of $5e-5$ is applied as a regularization term during training. We use the Cosine Annealing for learning rate scheduling, which gradually decreases the learning rate to $1e-5$. Our model is trained for 100 epochs on a single NVIDIA A100-SXM GPU. Our model is implemented in PyTorch 2.0.0.  As discussed in Section 3.4, the total loss is:
{\small
\begin{equation}\label{sup_eq:loss_total}
\mathcal L = \mathcal L_{sup}+ \mathcal L_{reg} +\alpha * \mathcal L_{sle}+\beta*\mathcal L_{c},
\end{equation}}

\noindent where $L_{sup}$ and $L_{reg}$ are basic losses from SMILE~\cite{ocd}. Following our definition, the $L_{sup}$  formulated as:
\begin{small}
\begin{equation}
    \mathcal{L}_{sup}=-\frac{1}{\left|P_i\right|} \sum_{p \in P_i} \log \frac{\exp \left(\mathcal{H}(f(\mathbf{x}_i)) \cdot \mathcal{H}(f(\mathbf{x}_p))\right)}{\sum_{j=1}^{|B|} \mathds{1}_{[j \neq i]} \exp \left(\mathcal{H}(f(\mathbf{x}_i)) \cdot \mathcal{H}(f(\mathbf{x}_j))\right)},
\end{equation}
\end{small}
where the $P_i$ is the positive set in a mini-batch and $|B|$ is batch size. Similarly, $L_{reg}$ is formulated as:
\begin{equation}
\hat{\boldsymbol{h}}_i=\operatorname{hash}\left(\mathcal{H}_{h}\left(f\left(\mathbf{x}_i\right)\right)\right),
\end{equation}

\begin{equation}
\mathcal{L}_{reg}=-\left|\hat{\boldsymbol{h}}_i\right|.
\end{equation}

\par\noindent
\textbf{Pseudo-code for Iterative Training}.
We iteratively perform leader feature generation and leader-based representation learning to dynamically update the leader features. The pseudo-code for the iterative training is elaborated below.

\begin{algorithm}[t]
\SetAlgoLined

\KwIn{Feature Extractor $f(\cdot)$, Projection Head $\mathcal{H}(\cdot)$, Labeled data $\mathcal{D}_{S}$ and synthesized data $\mathcal{D}^{G}$.}
\KwOut{$f$ and $\mathcal{H}(\cdot)$.}
\For{$n=1$ \textbf{in} $[1, 200]$}{

Extract features and execute clustering algorithm to assign category labels $\mathcal{Y}_{A}$\;
Generate a set of category-specific leader features $\mathcal{C}_{A}$\;

{
\For{$i=1$ \textbf{in} $[1, max\_iteration]$}{
Sample mini-batches from $\mathcal{D}_{S} \cup\mathcal{D}_{G}$\;
Calculate overall optimization objective by \cref{sup_eq:loss_total}\;
Update $f$ and $\mathcal{H}(\cdot)$ by SGD~\cite{qian1999momentum}\;
}

}
}

\caption{Iterative Training\label{sup_alg:iterative_training}}

\end{algorithm}

\begin{table}[t]
\small
 \setlength{\tabcolsep}{2pt}
\renewcommand{\arraystretch}{0.8}
\footnotesize
\centering

\caption{Results with Kmeans evaluation.\label{tab:upbound}} 

\begin{tabular}{c|ccc|ccc}
  \toprule
   \multirow{2}{*}{Method} & \multicolumn{3}{c|}{Mollusca}  & \multicolumn{3}{c}{CUB}\\
  \cmidrule(lr){2-4} \cmidrule(lr){5-7} 
   & All & Old & New  & All & Old & New \\
  \midrule
  Upbound &  39.4  &  44.6 &  36.7&  59.2 & 56.7 & 60.4
  \\ \midrule
Ours  & \textbf{35.2} & \textbf{43.3} & \textbf{30.8} & \textbf{54.1}  & \textbf{53.0}  & \textbf{54.7}
 \\
 SMILE  &29.5  & 34.1 & 27.1 & 49.8 & 46.2 & 51.6 \\ \bottomrule
 
\end{tabular}
\end{table}

\par\noindent
\textbf{Gradually Increase $\alpha$}.
As the OCD backbone network is unsupervised and pre-trained by the DINO~\cite{dino} approach that acquires a strong representation ability, model fine-tuning benefits from preserving the representation ability while adapting to target data. Inspired by the warm-up strategy~\cite{zheng2019pyramidal}, we design a linear growth paradigm for $\alpha$. From the $0^{th}$ to the $50^{th}$ epoch, the $\alpha$ increase from $0.1\times \alpha$ to $\alpha$. Then, the $\alpha$ is kept for the remaining 50 epochs.

\par\noindent
\textbf{Sampling Strategy of Leader-based Contrastive Learning}.
To facilitate the leader-based contrastive learning in a mini-bach, we follow~\cite{zheng2019pyramidal} to randomly sample $N^{C}\times N^{I}$ images to form a mini-batch, where $N^{C}$ is the number of categories in a mini-batch and $N^{I}$ is the number of images for each category. In our experiments, we set $N^{C}=8$ and $N^{I}=16$ for 128 batch size.

\section{Exploration Experiments}

\subsection{Gap in Data Synthesis.}
Although our DiffGRE can generate virtual images to help OCD model training, it is difficult to reasonably evaluate the quality of the generation. 
We conduct the exploration experiment in~\cref{tab:upbound}. We first use the $\mathcal{D}_{Q}$ as the generated data to execute our SLE training, and then directly use the Kmeans~\cite{kmeans} with ground-truth $K$ to generate category labels. We report the experimental results as ``Upbound'' in~\cref{tab:upbound}. Based on the results, we find that our virtual images contribute almost the same as real data on unknown categories. Moreover, compared with baseline accuracy based on SMILE, the gap between Upbound and ours is significantly narrowed by our DiffGRE framework, \eg, ALL-ACC is from 9.9\% to 4.2\% on the Mollusca dataset.

\subsection{Comparison with Diff-Mix}

\par\noindent
\textbf{Quantitative Comparison.}
In this section, we compare our method with the Diff-Mix data augmentation approach, which only augments known categories. We implement experiments on the CUB dataset. We followed the original settings of Diff-Mix and fed it with source images and targeted classes names to generate augmented images. Results are shown in Table~\ref{tab:compare_diffmix}. 
``SMILE + Diff-Mix'' refers to replacing the Attribute Composition Generation (ACG) module in the DiffGRE framework with Diff-Mix, in order to make a fair comparison.
Diff-Mix performs well on known categories, but there is no significant improvement on unknown categories. 
This proves that Diff-Mix can only synthesize images within known categories, leading to its inefficiency in the OCD task.
On the other hand, it is observed that our method achieves higher accuracy than Diff-Mix, which indicates the effectiveness of our implementation to synthesize novel samples belonging to unknown categories.

\begin{table}[t]
  \setlength{\tabcolsep}{8pt}
  \renewcommand{\arraystretch}{0.8}
  \footnotesize
  \begin{minipage}{0.48\textwidth}
    \centering
    \caption{Comparison with Diff-Mix on CUB dataset.}
    \begin{tabular}{l|ccc}
      \toprule
      \multirow{2}{*}{ Method } & \multicolumn{3}{c}{CUB } \\
      \cmidrule(lr){2-4} 
      & All & Old & New \\
      \midrule
SMILE~\cite{ocd} &  32.2 & 50.9 & 22.9   \\
      SMILE~\cite{ocd} + Diff-Mix~\cite{wang2024enhance} &  32.9 & 54.4 & 22.1   \\

     SMILE~\cite{ocd} + DiffGRE (Ours) & 35.4 & 58.2  & 23.8     \\
      \bottomrule
    \end{tabular}
    \label{tab:compare_diffmix}
  \end{minipage}

\end{table}

\begin{figure}[t]
  \centering
  \includegraphics[width=\linewidth]{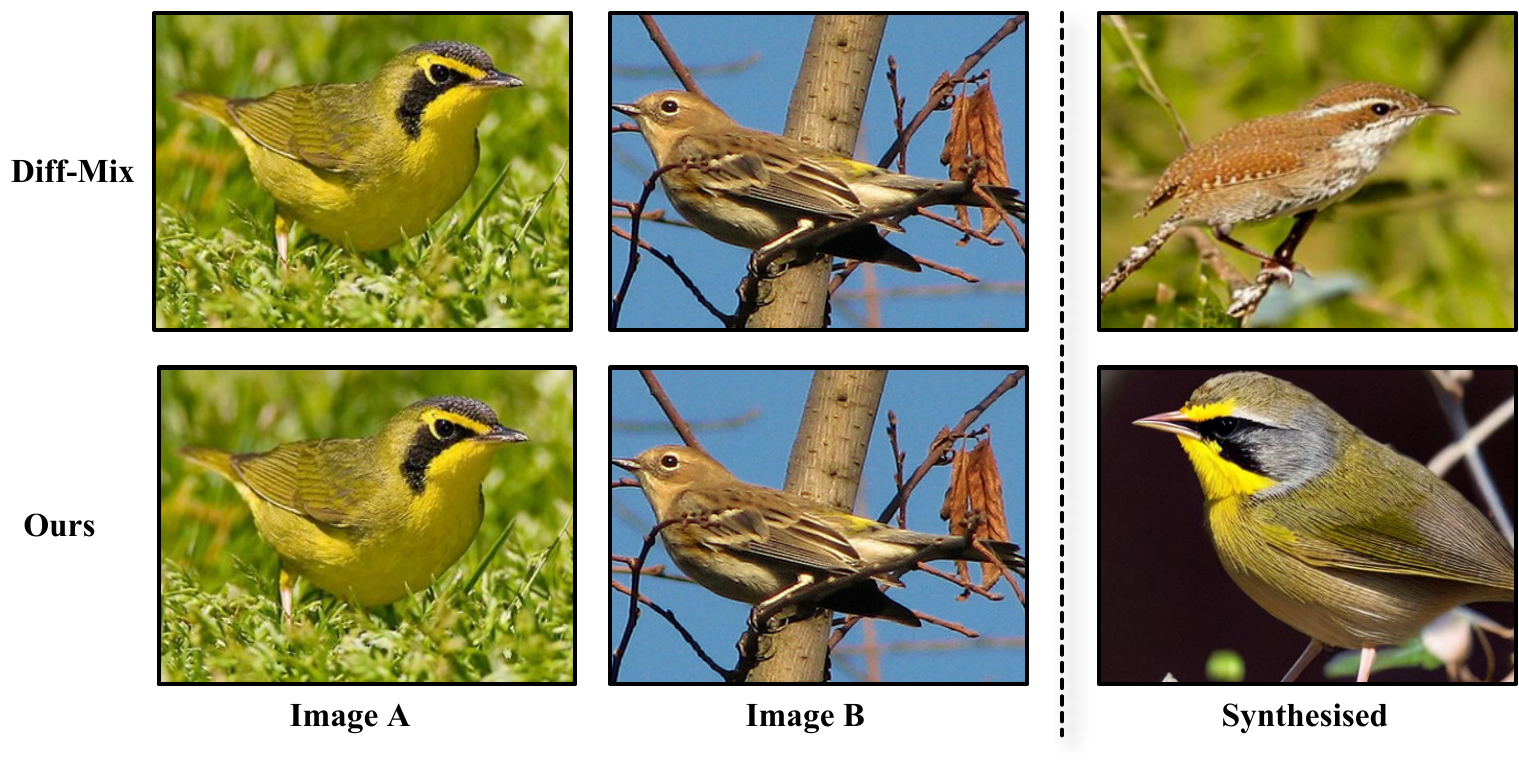}
  \caption{Our image generation compared with Diff-Mix~\cite{wang2024enhance} on the CUB dataset.\label{sup_fig:diffmix}}

\end{figure}

\begin{table}[t]
  \setlength{\tabcolsep}{3pt}
  \renewcommand{\arraystretch}{0.8}
  \footnotesize
  \begin{minipage}{0.48\textwidth}
    \centering
    \caption{Analysis on the hyperparameter $\gamma$.}
    \begin{tabular}{ccccc}
      \toprule

      \multirow{2}{*}{ Dataset } &  \multirow{2}{*}{ $\gamma$ } & Number of & Number of & Number of   \\
       &   & Remained  Images& Generated Images  & Labeled Images \\
      \midrule
      CUB & 0.40 &  1.2K  & 5.0K & 1.5K \\
      Scars & 0.65 & 1.1K  &  4.9K & 2.0K\\
     Pets &  0.25 & 0.3K &  1.4K & 0.9K \\
    Arachnida &  0.40 & 1.2K & 2.8K & 1.7K \\
    Animalia & 0.20  &  0.6K & 3.8K &1.5K \\
    Mollusca &  0.30 &  1.6K & 3.5K  & 2.4K \\
      \bottomrule
    \end{tabular}
    \label{tab:supp_gamma}
  \end{minipage}
\end{table}

\begin{table}[t]
\centering
\footnotesize
\setlength{\tabcolsep}{2pt}
\renewcommand{\arraystretch}{0.8}
\label{tab:statistics_images}
\caption{Statistics on SLE-based Clusters.\label{sup_tab:statistic}}
\begin{tabular}{l|ccc} 
    \toprule
    Data & $|\mathcal{Y}_A|$ & Img/ Cls & Img/ Cls (L)\\
    \midrule
    CUB & 116 &  23 & 15\\
    Scars & 171 & 27 & 20\\
     Pets & 37  & 19 &  49 \\
    Arachnida & 91 & 19 & 59 \\
    Animalia  &76 & 15  & 38 \\
    Mollusca & 103 & 28 & 51 \\
    \bottomrule
\end{tabular}
\end{table}

\par\noindent
\textbf{Qualitative Comparison.}
In addition to the quantitative comparison, we also conducted a qualitative comparison between our method and the Diff-Mix. Results are provided in Figure \ref{sup_fig:diffmix}. The Diff-Mix used an image from category $A$ as the source image and converted it to the targeted class $B$. The output is located in the top-right corner of Figure \ref{sup_fig:diffmix}.
Compared with Diff-Mix, our method takes two images from different categories as inputs and synthesizes a novel sample. It can be observed that our synthesized image is more likely to belong to an unknown category.

\begin{figure*}[t]
  \centering
  \includegraphics[width=\linewidth]{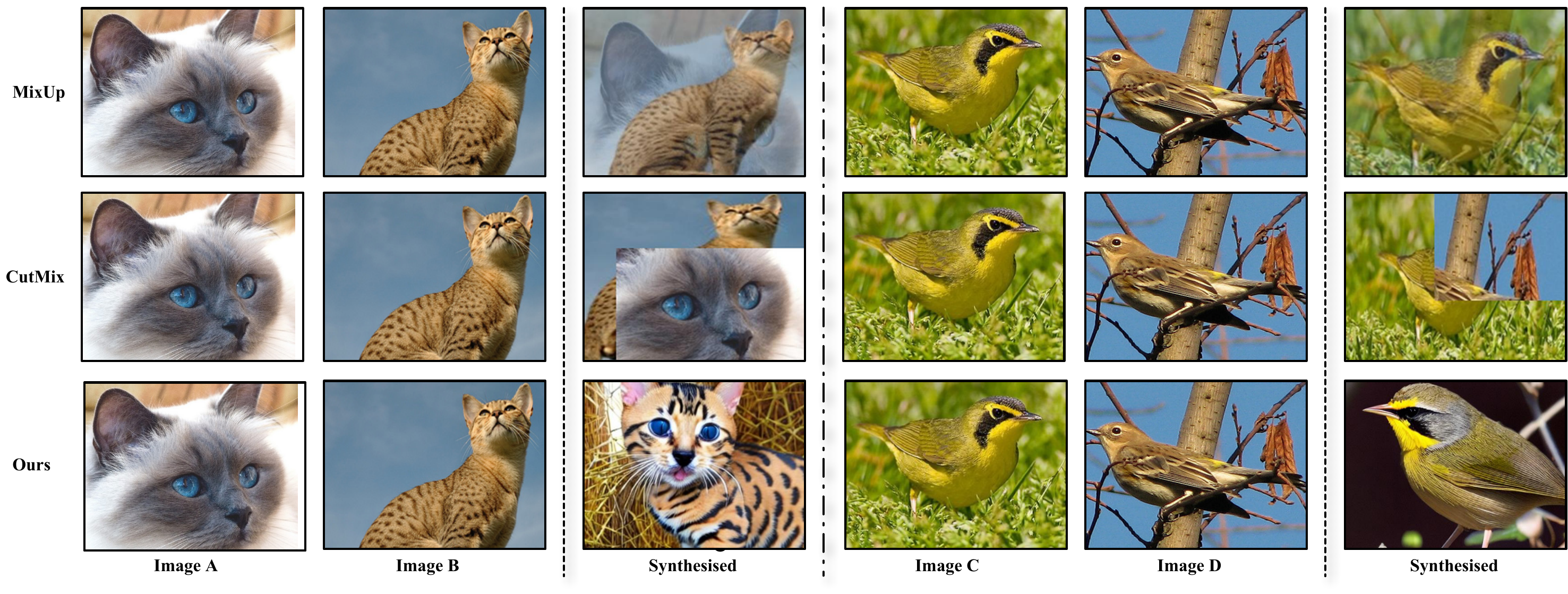}
  \caption{Our image generation compared with MixUp and CutMix on the CUB and the Pets dataset.\label{sup_fig:mixup_cutmix}}

\end{figure*}

\begin{table*}[t]
  \setlength{\tabcolsep}{10pt}
  \renewcommand{\arraystretch}{1.0}
  \footnotesize
  \begin{minipage}{1.0\textwidth}
    \centering
    \caption{Comparison with CutMix and MixUp on CUB, Stanfors Cars and Oxford Pets.}
    \begin{tabular}{l|ccccccccc}
      \toprule
      \multirow{2}{*}{ Method } & \multicolumn{3}{c}{CUB } & \multicolumn{3}{c}{ SCars} &\multicolumn{3}{c}{Pets} \\
      \cmidrule(lr){2-4} \cmidrule(lr){5-7} \cmidrule(lr){8-10}
      & All & Old & New & All & Old & New & All & Old & New\\
      \midrule
      SMILE~\cite{ocd} & 32.2 & 50.9 & 22.9 &  26.2 & 46.7 & 16.3  &  41.2  & 42.1 & 40.7    \\
      SMILE~\cite{ocd} + CutMix~\cite{yun2019cutmix} & 32.8 &  54.6 & 21.9
   &   28.2 & 55.0 &  15.3 &  40.9 & 46.3 & 38.0   \\

      SMILE~\cite{ocd} + MixUp~\cite{zhang2018mixup} & 33.3 & 55.6 &  22.1
 &  26.9 & 51.8 & 14.9
 & 37.9 & 41.6 & 35.9 \\

     SMILE~\cite{ocd} + DiffGRE (Ours) &  35.4 & 58.2  & 23.8 & 30.5 & 59.3 & 16.5 &  42.4 & 42.1 & 42.5  \\

      \bottomrule
    \end{tabular}
    \label{tab:compare_mix}
  \end{minipage}

\end{table*}

\subsection{Comparison with MixUp and CutMix}
\par\noindent
\textbf{Quantitative Comparison.} To better understand our method, we conducted experiments on three datasets CUB, Stanfors Cars and Ocford Pets to compare our method with two different mixing methods CutMix and MixUp. Quantitative results are summarized in Table~\ref{tab:compare_mix}. 
Similarly, ``SMILE + CutMix'' and ``SMILE + MixUp'' involve replacing the Attribute Composition Generation (ACG) module in the DiffGRE framework with CutMix and MixUp, respectively, to ensure a fair comparison.
We can observe that our method  significantly outperforms other methods. Especially in the unknown categories, our method achieves higher accuracy than CutMix and MixUp across all the three datasets. The results further demonstrate the effectiveness of dual interpolation in the latent space of the Diversity-Driven Refinement (DDR) module, compared to directly performing pixel-level interpolation.

\par\noindent
\textbf{Qualitative Comparison.} We also visualize the generated images on different datasets (CUB and Oxford Pets) from our method, MixUp and CutMix. Examples are illustrated in Figure~\ref{sup_fig:mixup_cutmix}. First,
we notice that for MixUp and CutMix, the results look like two images are randomly stitched together. But our synthesized images are more like real images. Second, our synthesized images simultaneously incorporate attributes from both input images. For example, in the case on the left, the synthesized cat has blue eyes, which is the same as the input image A. This visualization intuitively demonstrates the effectiveness of semantic latent interpolation in the Diversity-Driven Refinement (DDR) module.

\begin{figure*}[t]
  \centering
  \includegraphics[width=\linewidth]{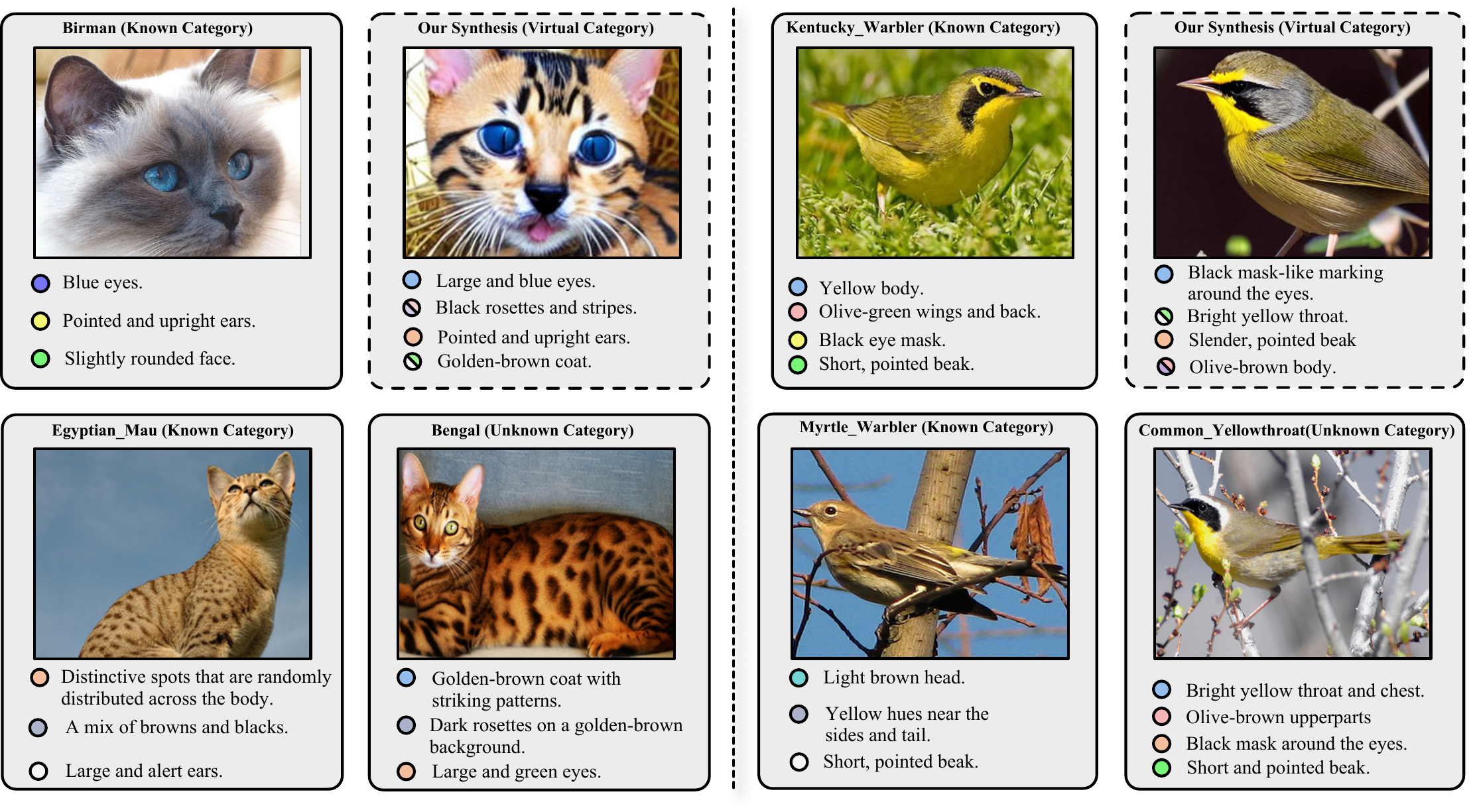}
  \caption{Additional Examples for Attribute Composition Generation (ACG).\label{sup_fig:visualization}}
\end{figure*}

\subsection{Details of Diversity-Driven Refinement}
\label{sup_sec:ddr}
In Section 4.4, we discussed the impact of $\gamma$ in the  Diversity-Driven Refinement (DDR) 
module on the Arachnida dataset, and it is observed that the model achieves the best performance on unknown categories when the number of the remaining images is comparable to that of labeled data. In this section, further details about the chosen value of $\gamma$, number of remained images and synthesized images are shown in Table~\ref{tab:supp_gamma}. Following the similar strategy, we set different values for $\gamma$ on six datasets in our experiments to ensure the number of remained images matches the size of the labeled training set.

On the other hand, we summarize the clustering results of SLE in ~\cref{sup_tab:statistic}, where the first column shows the number of categories to which samples from both known and virtual categories have been assigned. Moreover, we compared the average number of images per category (the 3rd Column) from SLE with the labeled training set (the 4th Column) and find they share the same scale. These results confirm sufficiency of the generated samples.

\begin{table}[t]
  \setlength{\tabcolsep}{8pt}
  \renewcommand{\arraystretch}{0.8}
  \footnotesize
  \begin{minipage}{0.48\textwidth}
    \centering
    \caption{Virtual Category Assignment Strategy}
    \begin{tabular}{c|l|ccc}
      \toprule
      \multirow{2}{*}{ Index }& \multirow{2}{*}{ Method } & \multicolumn{3}{c}{CUB } \\
      \cmidrule(lr){3-5} 
       & & All & Old & New \\
      \midrule
\textbf{a)}&$w/o$ class centers &  33.5 & 55.5 & 22.5    \\
   \textbf{b)} &$w/o$ category assignment & 31.3 & 51.9 & 20.9   \\

     \textbf{c)} & SMILE~\cite{ocd} + DiffGRE (Ours) & 35.4 & 58.2  & 23.8     \\
      \bottomrule
    \end{tabular}
    \label{tab:category_assignment}
  \end{minipage}
\vspace{-0.4cm}
\end{table}

\begin{table*}[t]
  \setlength{\tabcolsep}{4pt}
  \renewcommand{\arraystretch}{0.8}
  \footnotesize
  \begin{minipage}{1.0\textwidth}
    \centering

    \caption{Comparison of baseline methods on training time, inference time, model size, and ALL-ACC for the CUB dataset. ``*'' denotes results obtained with 4 GPUs in parallel.}
    \begin{tabular}{l|c|cc|cc|c|c}
      \toprule
      \multirow{2}{*}{ Method } & Inference  & Finetune Time & Synthesis Time  & Training Time & Model Size  & Inference Time  & ALL-ACC \\
       &  Strategy &  $\approx$(h) &  $\approx$(h) & $\approx$(s)& $\approx$(M) &  $\approx$(s) & (\%) \\
      \midrule
      SMILE~\cite{ocd} &  Hash-like& 0 &  0  & 960.6   &  13.4 & 11.2  & 32.2  \\
      SMILE~\cite{ocd} + Diff-Mix~\cite{wang2024enhance} & SLE-based & 16 &  2$^*$ &  8799.9   & 13.4  & 17.5  & 32.9 \\
     SMILE~\cite{ocd} + DiffGRE (Ours) & SLE-based & 0  & 12  &  8799.9    & 13.4  & 17.5  &  35.4 \\

     SMILE~\cite{ocd} + DiffGRE ($w/o$ DDR) &SLE-based & 0  & 12 & 9160.5     & 13.4  & 17.5  & 30.4  \\
      \bottomrule
    \end{tabular}
    \label{tab:computational}
  \end{minipage}
\end{table*}

\begin{table}[t]
  \setlength{\tabcolsep}{8pt}
  \renewcommand{\arraystretch}{0.8}
  \footnotesize
  \begin{minipage}{0.48\textwidth}
    \centering
    \caption{Evaluation on the hyperparameter $\beta$.}
    \begin{tabular}{ccccccc}
      \toprule
      \multirow{2}{*}{ Value } & \multicolumn{3}{c}{CUB } & \multicolumn{3}{c}{SCars}  \\
      \cmidrule(lr){2-4} \cmidrule(lr){5-7}
      & All & Old & New & All & Old & New \\
      \midrule
      0.5 & 34.6 &  57.6 & 23.0
 &  28.4 & 54.7 & 15.7
   \\
      0.75 &  33.9 & 56.2 & 22.8
 &  28.1 & 54.0 & 15.5
  \\
     1.0&   35.4 & 58.2 &23.8  &  30.5 & 59.3 & 16.5  \\
    1.25&  34.8 & 57.8 & 23.2
 & 26.9 & 49.9 & 15.7 \\
    1.5&  33.2 & 55.1 & 22.3
   & 27.4 & 51.6 & 15.7
 \\
      \bottomrule
    \end{tabular}
    \label{tab:hyper_beta}
  \end{minipage}
\vspace{-0.4cm}
\end{table}

\subsection{Virtual Category Assignment Strategy}

\begin{figure*}[t]
  \centering
  \includegraphics[width=\linewidth]{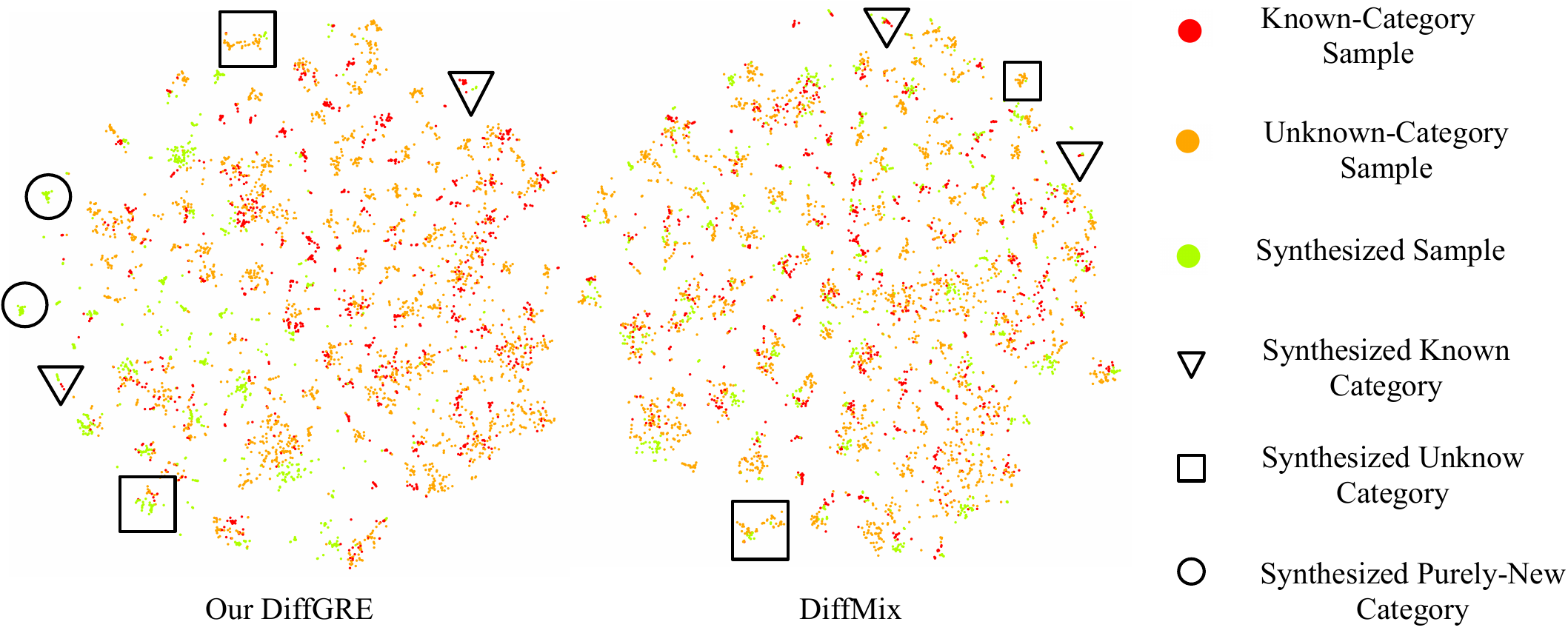}
  \caption{TSNE visualization of the feature of the images generated by our DiffGRE and Diff-Mix~\cite{wang2024enhance}. Through comparing these two sub-figures, we find that the images generated by Diff-Mix~\cite{wang2024enhance} are surrounded by the known-category and unknown-category samples, which indicates that the diversity of the generated images is limited. In contrast, our DiffGRE can generate diverse images that even belong to synthesized purely new categories. Thus, our methods are more effective in improving OCD performance by generating additional and virtual category information.\label{sup_fig:tsne}}
\end{figure*}

\begin{table*}[t]
  \caption{
  Comparison with \textbf{training-free hash-like inference} methods, with the best results in \textbf{bold} and the second best \underline{underlined}.}
  \setlength{\tabcolsep}{8pt}
  \renewcommand{\arraystretch}{1}
  \label{tab:training_free_hash_based}
  \centering
  \footnotesize 
  \begin{tabular}{l|ccc|ccc|ccc|ccc} 
    \toprule
    \multirow{2}{*}{Method} & \multicolumn{3}{c|}{\textbf{Arachnida}} & \multicolumn{3}{c|}{\textbf{Mollusca}} & \multicolumn{3}{c|}{\textbf{Oxford Pets}}  & \multicolumn{3}{c}{\textbf{Average}} \\
    \cline{2-13}
    & All & Old & New & All & Old & New & All & Old & New & All & Old & New \\

    \midrule

RankStat~\cite{rs} &26.6 &51.0 &10.0 & 29.3 & \underline{55.2} & 15.5 & 33.2 & \underline{42.3} &28.4& 29.7
  & 49.5
 &  18.0\\

WTA~\cite{wta} & \underline{28.1} & \underline{55.5} &10.9   &30.3 & \textbf{55.4} & 17.0 & 35.2 & \textbf{46.3} &29.3& 31.2  & \textbf{52.4} & 19.1  \\

SMILE~\cite{ocd} + DiffGRE (Ours) & \textbf{35.4} & \textbf{66.8} & \textbf{15.6} & \textbf{36.5} & 44.2& \textbf{32.5} & \textbf{42.4} & 42.1 & \textbf{42.5} & \textbf{38.2}  & \underline{51.0}    & \textbf{30.2} \\

    \bottomrule
    \multirow{2}{*}{Method} & \multicolumn{3}{c|}{\textbf{CUB}} & \multicolumn{3}{c|}{\textbf{Stanford Cars}} & \multicolumn{3}{c|}{\textbf{Animalia}} & \multicolumn{3}{c}{\textbf{Average}} \\
    \cline{2-13}
    & All & Old & New & All & Old & New & All & Old & New & All & Old & New \\
    \midrule

RankStat~\cite{rs} & 21.2 &26.9 &18.4 & 14.8 &19.9 & 12.3 &  31.4 & 54.9 & 21.6& 22.5 & 33.9  & 17.4   \\

WTA~\cite{wta}  &21.9 &26.9 &19.4  &17.1 & 24.4 & 13.6  & 33.4 &59.8 & 22.4& 24.1 & 37.0  & 18.5\\

SMILE~\cite{ocd} + DiffGRE (Ours) & \textbf{35.4} & \textbf{58.2} & \underline{23.8} & \textbf{30.5} & \textbf{59.3} & \underline{16.5} & \textbf{37.4} & \textbf{69.3} & 24.3 & \textbf{34.4}  & \textbf{62.3} &   \underline{21.5} \\

    \bottomrule
  \end{tabular}

  \vspace{-1.0em}
\end{table*}

We verify the effectiveness of our design for the virtual category assignment in~\cref{tab:category_assignment}. 
Specifically, \textbf{a)} by replacing class center distances with distances to all training samples, the results show a decrease in performance compared to our proposed design. \textbf{b)} When virtual category assignment is removed, performance further drops. \textbf{c)} These results are compared to the reference performance of our full method, which highlights the effectiveness of our approach.

\section{Additional Visualization Results}
\subsection{TSNE comparison between our DiffGRE and Diff-Mix}\label{sup_sec:tsne}
We visualize the features generated by our DiffGRE and  Diff-Mix~\cite{wang2024enhance} by TSNE in~\cref{sup_fig:tsne}.
Through comparing these two sub-figures, we find that the images generated by Diff-Mix~\cite{wang2024enhance} are surrounded by the known-category and unknown-category samples, which indicates that the diversity of the generated images is limited. In contrast, our DiffGRE can generate diverse images that even belong to synthesized purely new categories. Thus, our methods are more effective in improving OCD performance by generating additional and virtual category information.

\subsection{Additional successful examples}
We discussed visualization results in Section 4.5 in our main submission. In this section, we provide additional examples for Attribute Composition Generation (ACG). These examples are presented in Figure~\ref{sup_fig:visualization}. Positive examples can support
that our method is able to synthesize novel samples that include additional category knowledge. An example is shown on the right of Figure~\ref{sup_fig:visualization}. We find that the synthesized sample has attributes such as a black mask-like marking around the eyes, an olive-brown body, and a short, pointed beak, closely resembling a sample from the unknown category Common Yellowthroat.

\section{Computational Consumption}
\vspace{-0.5em}

We also compare computational costs of our framework, SMILE and Diff-Mix in Table~\ref{tab:computational},
analyzing Training Time, Inference Time, ALL-ACC, and other metrics.
Experiments are conducted on the CUB dataset. Diff-Mix is applied to generate 5000 images, matching the number of images generated by our framework.
``SMILE + Diff-Mix'' refers to substituting the Attribute Composition Generation (ACG) module in the DiffGRE framework with Diff-Mix, in order to provide a balanced comparison.
Notably, Diff-Mix involves two steps for generating new images: first, finetuning its diffusion model, and second, using the pre-trained model to synthesize samples. 
Our framework achieves better performance with slightly higher inference time than SMILE.
On the other hand, 
the total generation time for ``SMILE + Diff-Mix'' is $\textbf{16} + \textbf{2} = \textbf{18}$ hours, while our framework completes the process in just \textbf{12} hours, as it does not require finetuning.
Additionally, when the DDR module is removed from our framework, we observe longer training time and lower accuracy, further highlighting the importance of the Diversity-Driven Refinement (DDR) module.

\section{Additional Hyper-Parameter Analyses}
\vspace{-0.5em}
\par
\begin{figure}[t]
  \centering
  \includegraphics[width=\linewidth]{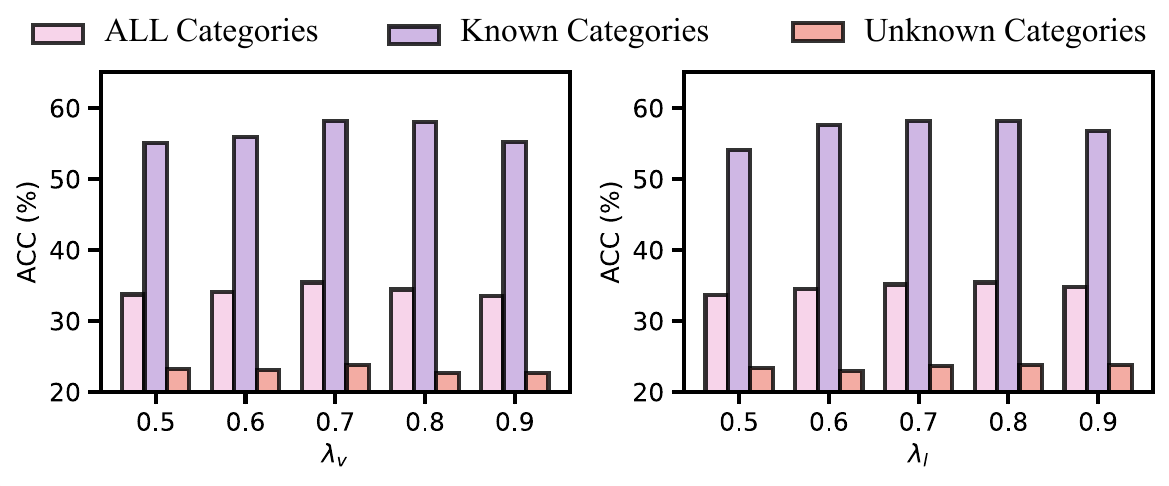}
  \caption{Illustration for the impact of the varying interpolation parameters, $\lambda_{v}$ and $\lambda_{l}$, for visual embedding and latent embedding, respectively.\label{sup_fig:lambda2}}
  \vspace{-1.5em}
\end{figure}

\noindent\textbf{Interpolation Parameters.}
As shown in~\cref{sup_fig:lambda2}, we analyzed the impact of $\lambda_{v}$ and $\lambda_{l}$, for visual embedding interpolation and latent embedding interpolation. Theoretically, if $\lambda_v$ or $\lambda_l$ are close to 0/1, the generated image resembles one of the original images, resulting in low diversity. At 0.5, there is higher diversity but also increased ambiguity. The results show that the parameters are not sensitive, thus we empirically set these interpolation parameters to $\lambda_{v}=0.7 $ and $\lambda_{l}=0.8$, respectively.

\par\noindent\textbf{Loss Weight.} We analyzed the impact of hyperparameter $\beta$ in Section 4.4. Additional results of the analysis on $\beta$ are summarized in Table~\ref{tab:hyper_beta}. Experiments are implemented on two datasets CUB and Stanford Cars. Our model attains the highest accuracy when $\beta$ increases from 0.5 to 1.5 on the two datasets. Based on the results, we set $\beta=1.0$ for all datasets during training.

\section{Additional Evaluation of our DiffGRE compared with training-free hash-like methods} 

We further conduct experiments to evaluate our method compared with training-free hash-like approaches. Results are provided in \cref{tab:training_free_hash_based}. The combination of ``SMILE + DiffGRE'' consistently outperforms all competitors, achieving the highest ACC-ALL averages, which surpasses WTA by 8.7\% and RankStat by 10.2\% across the six datasets.

\section{Limitation} Our framework includes the offline generation. While this reduces the computational cost, the diffusion model lacks end-to-end optimization, thereby leading to unsatisfactory generative results. We will explore finetune-based approaches in the future.

\end{document}